\documentclass[10pt,twocolumn,letterpaper]{article}

\usepackage{cvpr}              
\definecolor{cvprblue}{rgb}{0.21,0.49,0.74}
\usepackage[pagebackref,breaklinks,colorlinks,allcolors=cvprblue]{hyperref}

\usepackage{amsthm}
\theoremstyle{definition}
\newtheorem{proposition}{Proposition}
\usepackage{enumitem}
\newlist{propenum}{enumerate}{1}
\setlist[propenum,1]{label=(\roman*), leftmargin=2em, nosep}

\def\paperID{41370} 
\def\confName{CVPR}
\def\confYear{2026}

\title{ACPV-Net: All-Class Polygonal Vectorization for Seamless Vector Map Generation from Aerial Imagery}

\author{
Weiqin Jiao$^{*}$ \quad Hao Cheng \quad George Vosselman \quad Claudio Persello \\
Faculty of Geo-Information Science and Earth Observation (ITC)\\
University of Twente, The Netherlands\\
{\tt\small w.jiao@utwente.nl}
}

\usepackage{multirow} 
\usepackage[dvipsnames,table]{xcolor} 
\definecolor{confblue}{RGB}{20,44,122} 
\definecolor{oursbg}{HTML}{FFF4D6} 
\usepackage{comment}

\newcommand{\conf}[2]{\textit{\textcolor{confblue}{\footnotesize(#1~'#2)}}}
\usepackage[ruled,vlined]{algorithm2e} 
\usepackage{amssymb}
\usepackage{pifont} 

\usepackage[accsupp]{axessibility}  

\usepackage[x11names,dvipsnames]{xcolor}

\begin{document}
\twocolumn[{%
\renewcommand\twocolumn[1][]{#1}%
\vspace{-13mm}
\maketitle
\vspace{-11mm}
\begin{center}
    \captionsetup{type=figure}
    \includegraphics[width=\linewidth,trim={58mm 44mm 58mm 46mm},clip]{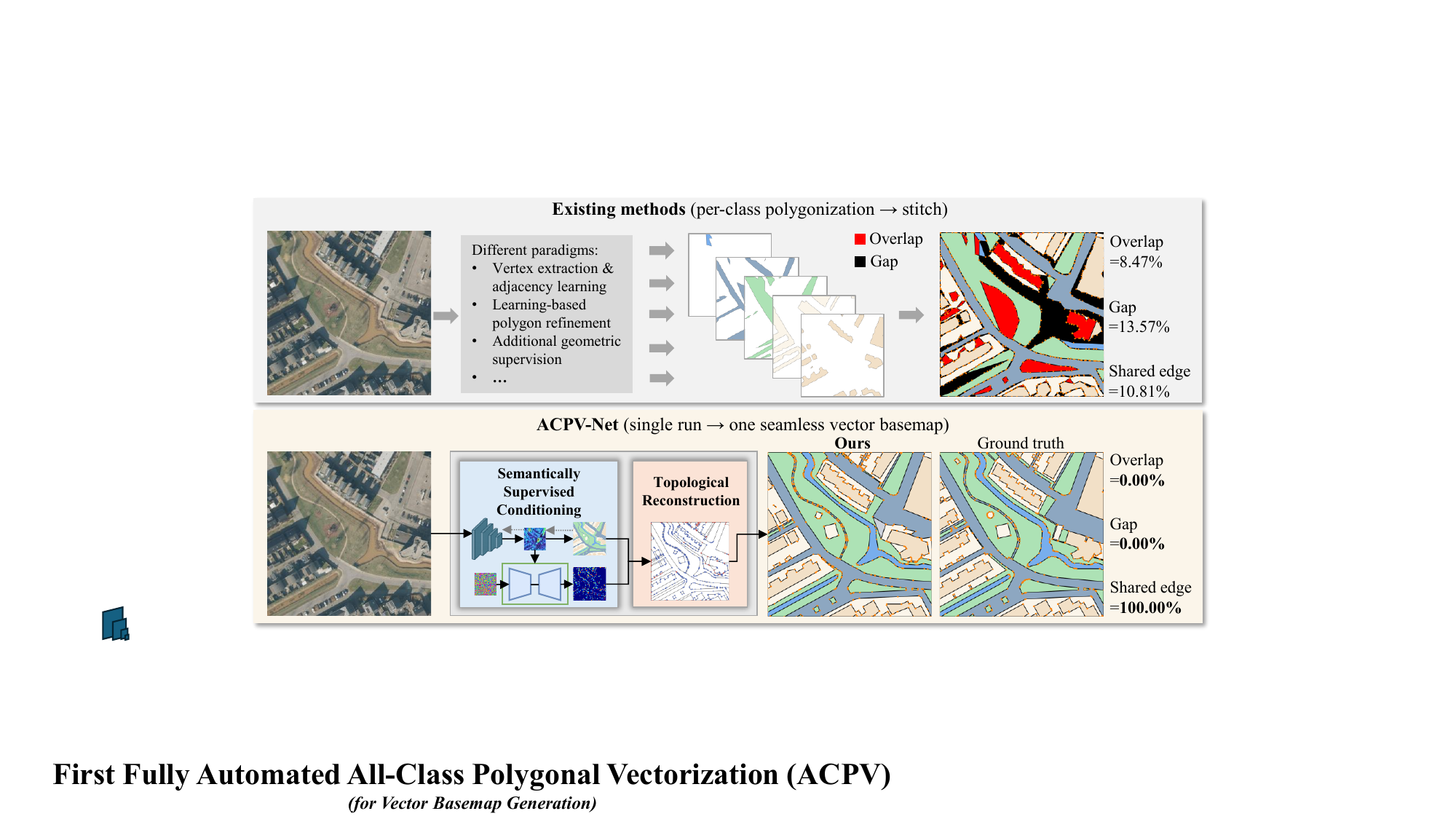}
  \caption{To generate an all-class vector basemap, existing single-class polygonization methods require per-class inference and stitching, resulting in gaps and overlaps across classes. \textbf{ACPV-Net} is the \textbf{first fully automatic framework} that produces a seamless basemap with \textbf{shared-edge consistency} and \textbf{avoids gaps and overlaps} in a single run.}
  \label{fig:teaser}
\end{center}%
}]

\begin{abstract}
We tackle the problem of generating a complete vector map representation from aerial imagery in a single run: producing polygons for all land-cover classes with shared boundaries and without gaps or overlaps. 
Existing polygonization methods are typically class-specific; extending them to multiple classes via per-class runs commonly leads to topological inconsistencies, such as duplicated edges, gaps, and overlaps. 
We formalize this new task as All-Class Polygonal Vectorization (ACPV) and release the first public benchmark, Deventer-512, with standardized metrics jointly evaluating semantic fidelity, geometric accuracy, vertex efficiency, per-class topological fidelity and global topological consistency. 
To realize ACPV, we propose ACPV-Net, a unified framework introducing a novel Semantically Supervised Conditioning (SSC) mechanism coupling semantic perception with geometric primitive generation, along with a topological reconstruction that enforces shared-edge consistency by design. 
While enforcing such strict topological constraints, ACPV-Net surpasses all class-specific baselines in polygon quality across classes on Deventer-512.
It also applies to single-class polygonal vectorization without any architectural modification, achieving the best-reported results on WHU-Building. 
Data, code, and models will be released at: \url{https://github.com/HeinzJiao/ACPV-Net}.
\end{abstract}    
\section{Introduction}
\label{sec:intro}
A vector basemap, a.k.a. a topographic map, is a seamless, multi-class, vector representation of land cover, where each region is encoded as a polygon of a specific class, and adjacent polygons share precisely one common boundary without gaps or overlaps \cite{ArcGISProTopologyPolygons, OGC_SFA_Part1, ISO19125_1_2004, kent2009topographic, hohle2017generating, kent2018topographic, elberink2010acquisition, GegevenscatalogusBGT}. 
Such basemaps form the foundation of national geospatial data infrastructures and are essential for cadastral management and land-use planning. 
Despite their importance, the production and maintenance of authoritative vector basemaps remain predominantly manual \cite{EsriVectorBasemapsOverview,GoogleMaps101_2022,GoogleVectorMapDocs}, leading to labor-intensive, costly, and poorly reproducible workflows. 

Learning-based polygonization methods \cite{zhao2021building,girard2021polygonal,zorzi2022polyworld,xu2023hisup,jiao2024polyr,jiao2025roipoly,yang2023topdig,zhang2025global,jiao2025ldpoly} have advanced automated map vectorization but remain class-specific. 
They are typically run per class and then stitched to obtain a complete vector basemap. 
Such post-hoc stitching commonly breaks topology, leading to gaps, overlaps, and duplicated edges (Fig.~\ref{fig:teaser}), and thus falls short of the shared-edge and zero-gap requirements. 

In this paper, we formalize this new task as \emph{All-Class Polygonal Vectorization (ACPV)}: generating a single, globally consistent \emph{planar partition} of the image domain into per-class polygons that satisfy strict topological constraints (formal definitions in Sec.~\ref{sec:acpv}). 
Compared with single-class polygon extraction, ACPV is substantially harder, as it requires unified reasoning over semantics and geometry under a globally consistent topological structure. 
Here, geometry refers to the geometric primitives that define polygonal structures, \eg, edges, vertices, or equivalent structures, that encode region boundaries and support topological reconstruction. 
This challenge arises from five key factors: 
(i) \textbf{Semantic–geometric heterogeneity} (raster, categorical semantics vs.\ vector, continuous geometry) complicates joint optimization; 
(ii) \textbf{Strict alignment} is needed between semantic regions and geometric boundaries to preserve topological consistency; 
(iii) \textbf{Weak/ambiguous visual cues} in aerial imagery (shadows, occlusions, semantically ambiguous boundaries) demand reasoning beyond appearance; 
(iv) \textbf{Cartographic conventions} (heuristic rules on how densely to sample or simplify vertices along smooth boundaries) are difficult to encode explicitly in standard learning frameworks; 
and (v) \textbf{Global topological reconstruction} of a single planar partition with shared boundaries and no gaps/overlaps goes beyond per-class geometry. 

To address these challenges, we introduce \textbf{ACPV-Net}, the first fully automatic framework that converts a single aerial image into a topologically consistent vector basemap, making a significant step toward fully automatic topographic mapping. 
We model geometry as discrete polygonal vertices, 
the atomic elements of polygonal structures, 
as Gaussian-mixture vertex heatmaps, allowing semantics and geometry to share the same raster domain. 
This distributional vertex representation enables the diffusion model to (1) learn cartographic sampling conventions that reflect human mapping heuristics,
and (2) infer sparse, sharp vertex peaks under weak/ambiguous image cues such as shadows, occlusions, or semantically ambiguous boundaries.
This generative, task-coupled geometry extraction stands in contrast to existing polygonal outline extraction methods \cite{zorzi2022polyworld,xu2023hisup,yang2023topdig,zhang2025global}, which lack learned priors for cartographic conventions and degrade under weak image cues.

Our core mechanism, \emph{Semantically Supervised Conditioning (SSC)}, directly supervises the diffusion conditioning stream with a semantic segmentation loss, enabling the conditioning itself to learn task-coupled semantics. 
Unlike existing conditional diffusion pipelines \cite{rombach2022high,zhang2023adding,yang2024pixel,zhao2025local} that inject external conditions without explicit supervision, SSC turns the conditioning into an active guidance signal that enforces semantic–geometric alignment and focuses vertex generation within class-consistent regions. 
The coherent evidence (multi-class mask + vertex peaks) completes the ACPV formulation through a proposition-driven Planar Straight-Line Graph (PSLG) reconstruction, which provably ensures topological consistency by design rather than through heuristic post-processing (see Sec.~\ref{sec:acpv},~\ref{sec:topo_recon} for formal proof). 

To evaluate this new task and framework, we establish \textbf{Deventer-512}, the first public benchmark for ACPV. 
Existing datasets either target single-class polygon extraction or provide multi-class raster masks without vector geometry, leaving no standard benchmark for ACPV. 
Deventer-512 fills this gap with high-resolution aerial imagery (0.3 m GSD) and fine-grained polygon annotations covering urban, suburban, and rural scenes under challenging conditions (shadows, occlusions, and semantically ambiguous boundaries). 
It is representative yet compact and computationally accessible ($\sim$2k patches), comprising 1{,}716 training, 212 validation, and 220 testing patches, totaling 84{,}403 instances and 22{,}679 internal holes (the most complex polygon contains up to 708 vertices), covering five typical land-cover classes with diverse geometric complexity, suitable for assessing both semantic and geometric reasoning. 
A standardized evaluation protocol jointly measures semantic fidelity, geometric accuracy, vertex efficiency, per-class topological fidelity and global topological consistency. 
We also provide subsets for the above challenging conditions to enable robustness evaluation under weak/ambiguous evidence. 

Our main contributions are summarized as follows: 
\begin{itemize}
\item We address the vector basemap generation problem by formulating it as \emph{All-Class Polygonal Vectorization (ACPV)}: generating a single, globally consistent \emph{planar partition} of the image domain into per-class polygons that satisfy strict topological consistency. 
\item We introduce \textbf{ACPV-Net}, the first fully automatic framework that, in a single run, produces topologically consistent vector basemaps from aerial imagery.
\item We release \textbf{Deventer-512}, the first public ACPV benchmark with high-resolution imagery, fine-grained polygons across five land-cover classes, a standardized evaluation protocol, and subsets for challenging conditions. 
\item ACPV-Net achieves gap-/overlap-free topology and significantly outperforms strong class-specific baselines across land-cover classes on Deventer-512. It also applies, without any architectural change, to single-class polygonal vectorization on the WHU-Building dataset while maintaining excellent performance. 
\end{itemize}

\section{Related work}
\label{sec:related_work}
\noindent\textbf{Single–class polygonal outline extraction.}
Most learning-based methods \cite{zorzi2022polyworld,xu2023hisup,yang2023topdig,zhang2025global,adimoolam2025pix2poly,girard2021polygonal} address a single class, typically buildings, given their practical importance in mapping and the abundance of high-quality vector annotations. 
To construct a complete all-class map, their per-class results must be stitched post hoc, often introducing gaps, overlaps, or duplicated boundaries. 
Representative paradigms include learning-based polygon generation/refinement (\eg, \cite{peng2020deep, zhang2025global}), vertex detection and adjacency learning (\eg, \cite{yang2023topdig, zorzi2022polyworld, zorzi2023re, adimoolam2025pix2poly}), and additional geometric supervision such as attraction or frame fields (\eg, \cite{girard2021polygonal, xu2023hisup}). 
RoI-based variants \cite{zhao2021building,jiao2024polyr,jiao2025roipoly} avoid heavy heads but still process cropped instances independently, hindering shared-edge reasoning and limiting scalability to area-covering or elongated classes. 
Recent works exploring class-agnostic or multi-class settings \cite{yang2023topdig,meng2025irsamap,wang2025pcp} remain primarily single-class in training or evaluation and combine per-class boundaries post hoc, without explicitly enforcing global topological consistency. 
In contrast, we pursue single-run, all-class polygonal vectorization that directly produces a topologically consistent planar partition of the image domain. 

\vspace{2pt}
\noindent\textbf{Segmentation-to-vectorization cascades.}
Multi-class land-cover mapping and semantic segmentation~\cite{wang2020deep, wang2021loveda, wang2022unetformer, li2023large, liu2024vmamba, zhu2024samba, li2025lsknet} output rasterized multi-class masks. 
Obtaining vectors often relies on geometry-driven, post-hoc simplifications such as Douglas–Peucker \cite{douglas1973algorithms}, applied per connected component of the mask. 
This often introduces duplicated or misaligned inter-class boundaries, gaps/overlaps, and redundant vertices, violating the formal constraints of a seamless vector basemap (specified in Sec.~\ref{sec:acpv}). 

\vspace{2pt}
\noindent\textbf{Conditional diffusion for structured geometry.}
Existing polygonization networks \cite{zorzi2022polyworld,xu2023hisup,yang2023topdig,zhang2025global,adimoolam2025pix2poly,girard2021polygonal} are predominantly \emph{discriminative}, rely on discriminative visual cues to detect geometric structures such as vertices. 
Such formulations often struggle under weak/ambiguous cues (\eg, shadows, occlusions, semantically ambiguous boundaries) and lack learned cartographic conventions. 
To overcome these limitations, diffusion-based generative models have shown potential to offer probabilistic reasoning that complements discriminative cues \cite{lin2024stable, jiao2025ldpoly}. 
However, existing conditional diffusion pipelines \cite{rombach2022high, zhang2023adding, yang2024pixel, zhao2025local} are primarily designed for generic visual synthesis, where external conditions (\eg, images, texts, or sketches) are effective but lack explicit semantic supervision for structured geometry or semantic alignment. 
We fill this gap by introducing Semantically Supervised Conditioning (SSC), which directly supervises the conditioning branch with a semantic loss to guide vertex generation. 
This enables the model to learn cartographic conventions and maintain semantically aligned, sharp vertex peaks under weak image cues (see Sec.~\ref{sec:ablation}). 

\vspace{2pt}
\noindent\textbf{Topology and planar graph reconstruction.}
Topological consistency is a fundamental requirement for seamless vector basemap generation, as formalized in cartographic and GIS standards that define planar partitions with shared boundaries and no gaps or overlaps \cite{ArcGISProTopologyPolygons,OGC_SFA_Part1,ISO19125_1_2004, utwente_ltb_concept81728}. 
Although some polygonal outline extraction methods 
\cite{zorzi2022polyworld,yang2023topdig,adimoolam2025pix2poly} incorporate graph-based designs to model vertex connectivity, and a few recent works (\eg \cite{chen2024automatic,xia2024vectorizing}) introduce heuristic post-processing or topology-aware losses to alleviate local errors, they focus on local topological relationships rather than ACPV's global topological consistency. 
To overcome this limitation, we introduce a proposition-driven planar straight-line graph (PSLG) reconstruction that ensures topological consistency, providing a globally consistent planar partition by construction (Sec.~\ref{sec:topo_recon}).

\section{The ACPV task}
\label{sec:acpv}
We define ACPV (All-Class Polygonal Vectorization) as the task of producing a topology-consistent polygonal partition for all land-cover classes from an aerial image. 

\noindent\textbf{Problem Definition}
Let $I:\Omega \to \mathbb{R}^k$ be an aerial image with $k$ spectral channels ($k=3$ for RGB) and $\mathcal{C}$ be a finite land-cover class set with $|\mathcal{C}|=C$, ACPV seeks a labeled polygonal partition 
$\{\mathcal{P}_c\}_{c\in\mathcal{C}}$ of the spatial domain $\Omega$ that satisfies the following constraints:

\noindent\textbf{(a) Planar partition.}
Polygon interiors are disjoint, and the union of all polygons fills the whole domain $\Omega$:
\begin{equation}
\begin{aligned}
\mathrm{int}(p_i)\cap\mathrm{int}(p_j) &= \varnothing \quad \text{ } \forall i\neq j,\\
\bigcup_{c}\ \bigcup_{p\in\mathcal{P}_c}\ \mathrm{int}(p) &= \Omega.
\end{aligned}
\end{equation}
where
$\mathcal{P}_c$ denotes the set of polygons labeled by class $c$ and $\mathrm{int}(p)$ is the interior of polygon $p$.

\noindent\textbf{(b) Shared boundaries.}
Any two adjacent polygons share one identical boundary segment, 
and no duplicate or parallel copies of the same boundary exist.

\noindent\textbf{(c) Zero gap/overlap.}
Each point in $\Omega$ belongs to exactly one polygon, except along shared boundaries.

\noindent\textbf{(d) Linear geometry.}
Each polygon boundary is a finite union of simple piecewise-linear chains (outer ring plus optional holes), without self-intersections.

\noindent\textbf{(e) Semantic consistency.}
Each polygon carries exactly one label $c\in\mathcal{C}$, and label transitions occur only across shared boundaries.

\noindent\textbf{(f) Minimal vertex redundancy.}
Each polygon boundary is represented in a canonical, vertex-minimal form: 
no duplicate consecutive vertices or geometrically redundant points are allowed along any linear segment; vertex reduction must preserve the polygon’s geometry and topology. 

\section{Method}
\label{sec:method}
\begin{figure*}[t]
  \centering
  \includegraphics[width=\linewidth,trim={5mm 105mm 5mm 0mm},clip]{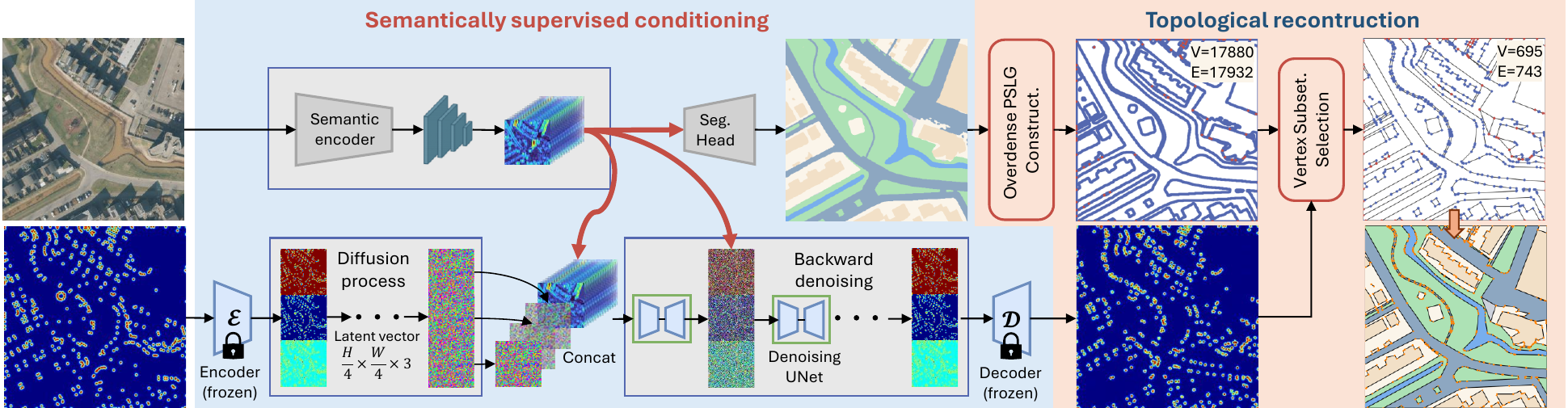}
  \caption{Overview of ACPV-Net. It unifies semantically supervised conditioning and proposition-driven topological reconstruction: the former produces coherent semantic–geometric evidence through diffusion-based vertex generation under semantic supervision, the latter deterministically reconstructs a topology-consistent vector basemap via overdense PSLG construction and vertex-guided subset selection.}
  \label{fig:network}
  \vspace{-6pt}
\end{figure*}

We aim to generate a seamless, topology-consistent vector basemap from a single aerial image $I$ -- a single planar partition of the image domain into per-class polygons satisfying the ACPV constraints (Sec.~\ref{sec:acpv}). 
We propose \textbf{ACPV-Net}, as shown in Fig.~\ref{fig:network}, a \emph{unified} framework with two tightly coupled components:

\begin{itemize}
\item \textbf{Semantically supervised conditioning (SSC).} 
We choose vertices as the geometric primitives for constructing the vector basemap. 
Vertices of an aerial image are encoded as a \emph{Gaussian-mixture heatmap} and reconstructed in latent space with a diffusion model. 
The conditioning stream is \emph{explicitly} supervised by a semantic segmentation loss, so that semantics guide vertex generation and enforce semantic–geometric alignment.

\item \textbf{Proposition-driven topological reconstruction.} 
From the coherent evidence $(\hat{M},\hat{Y})$ with semantic masks $\hat{M}$ and vertex heatmap $\hat{Y}$, we reconstruct polygons through a \emph{planar straight-line graph (PSLG)}–based algorithm derived from a sufficient condition that \emph{ensures} global topological consistency by design.
\end{itemize}

\subsection{Semantically supervised conditioning}
\label{sec:ssc}
\noindent\textbf{Distributional vertex modeling.}
Let $y\!\in\![0,1]^{H\times W}$ be a Gaussian-mixture vertex heatmap (peaks centered at annotated corners). 
We encode $y$ into a latent $z_0=\mathcal{E}(y)\in\mathbb{R}^{C_z\times H'\times W'}$ ($H'=H/4$, $W'=W/4$) using a pretrained variational autoencoder (VAE) from ~\cite{rombach2022high}. 
As verified in the supplementary, the pretrained VAE preserves vertex-peak information with negligible degradation, making fine-tuning unnecessary; thus, it is kept frozen during training. 

\vspace{2pt}
\noindent\textbf{Task-coupled conditioning.}
A semantic encoder $S_\psi(I)\in\mathbb{R}^{C_s\times H'\times W'}$ provides conditioning features at the same spatial scale as $z_0$. Unlike generic conditional pipelines \cite{rombach2022high, zhang2023adding, yang2024pixel, zhao2025local}, we \emph{explicitly} supervise $S_\psi$ with a segmentation loss via a lightweight segmentation head, so that the conditioning carries discriminative semantics aligned with the downstream vertex generation.

\vspace{2pt}
\noindent\textbf{Denoising objective.}
With forward noising $q(z_t\!\mid\!z_{t-1})$ and schedule $\{\beta_t\}_{t=1}^{T}$, the latent diffusion model predicts the clean latent $\hat{z}_0$ from the noisy input $z_t$ conditioned on semantic features $S_\psi(I)$:
\begin{equation}
\hat{z}_0=\phi_\theta\big(z_t, t, \mathrm{Cond}(S_\psi(I), z_t)\big).
\end{equation}
where $S_\psi$ is trained \emph{jointly} with the denoiser under the unified loss: 
\begin{equation}
\resizebox{0.45\textwidth}{!}{%
$\displaystyle
\mathcal{L}_{\mathrm{SSC}}=
\lambda_{\epsilon}\,\mathbb{E}\|\epsilon-\epsilon_\theta(\cdot)\|_1+
\lambda_{0}\,\mathbb{E}\|z_0-\hat{z}_0\|_1+
\lambda_{\mathrm{seg}}\,\mathcal{L}_{\mathrm{seg}}(\hat{M},M),
\label{eq:loss-ssc}
$}
\end{equation}
where $\mathcal{L}_{\mathrm{seg}}$ is applied to the semantic prediction $\hat{M}$. 
At inference, iterative denoising yields $\hat{z}_0$, which is decoded to a vertex heatmap $\hat{Y}=\mathcal{D}(\hat{z}_0)$. 
The pair $(\hat{M},\hat{Y})$ constitutes coherent semantic–geometric evidence, with vertices spatially aligned with class boundaries in the semantic mask and covering key geometric corners. 
This coherent evidence serves as the input for the proposition-driven topological reconstruction detailed in Sec.~\ref{sec:topo_recon}.

\subsection{Proposition-driven topological reconstruction}
\label{sec:topo_recon}
We reconstruct a single planar partition from $(\hat{M},\hat{Y})$. 
We first formalize a \emph{sufficient condition} under which the reconstructed planar partition satisfies all ACPV constraints (a)–(f) (see Sec.~\ref{sec:acpv}); 
we then derive a \emph{deterministic} algorithm that satisfies this condition by construction. 
An overview of the process is shown in Fig.~\ref{fig:network}.

\begin{proposition}[Sufficient condition for ACPV compliance]
\label{prop:acpv-sufficient}
Let $G=(V,E)$ be a \emph{planar straight-line graph} embedded in $\mathbb{R}^2$. Suppose that:
\begin{propenum}
    \item Each edge $e \in E$ lies along a label-transition boundary in the multi-class mask $\hat{M}$, and the two faces adjacent to $e$ have distinct semantic labels; and
    \label{con1}
    \item The vertex set $V$ consists exclusively of \emph{geometric keypoints}, that is, vertices whose removal would alter the polygon’s topology or geometric structure. 
    \label{con2}
\end{propenum}
Then the polygonal partition obtained by tracing all face boundaries on $G$ and assigning face labels according to $\hat{M}$ satisfies all ACPV constraints (a)–(f). A complete proof is provided in the supplementary.
\end{proposition} 

\noindent\textbf{Deterministic algorithm.}
We enforce the two premises in Proposition~\ref{prop:acpv-sufficient} by construction: 

\noindent\textbf{(1) Overdense PSLG construction.}
From the predicted multi-class mask $\hat{M}$, we extract all label-transition loci on the pixel-center lattice (including image borders) and connect adjacent transition pixels into one-pixel-wide boundary chains. 
Each transition pixel is treated as a vertex, and adjacent pairs form edges, resulting in a lattice-aligned planar straight-line graph $G=(V,E)$. 
We deliberately keep all transition locations without any thinning or pruning, making $G$ \emph{globally overdense}, a superset covering all admissible boundary positions required by Proposition~\ref{prop:acpv-sufficient} Condition \ref{con1}. 
This step establishes the planar scaffold on which the subsequent vertex-guided selection operates to satisfy Proposition \ref{prop:acpv-sufficient} Condition \ref{con2}. 

\noindent (2) \textbf{Vertex-guided subset selection.}
We decompose the overdense PSLG $G=(V,E)$ at anchor vertices ($\deg(v)\!\neq\!2$) into anchor-bounded polylines. 
Discrete vertex peaks $\hat{V}_p$ extracted from $\hat{Y}$ are projected onto their nearest PSLG loci within a fixed radius~$\tau$ 
and ordered by arc length. 
The anchors and the projected keypoints are preserved to form simplified polylines that retain boundary geometry while removing redundant vertices. 
This step is designed to realize Proposition \ref{prop:acpv-sufficient} Condition \ref{con2}. 
Together with Step (1), which satisfies Condition \ref{con1}, it yields a simplified PSLG. 
Tracing the faces of this PSLG and assigning labels from $\hat{M}$ produces the final vectorized map. 

All operations are deterministic and use only protocol-level constants (\eg~$\tau$), without dataset-specific tuning. 

\section{The ACPV Benchmark}
\label{dataset}
Automated polygonal outline extraction has long relied on single-class datasets (\eg, Inria~\cite{maggiori2017can}, WHU-Building~\cite{ji2018fully}), 
where only one class is vectorized and topology across land-cover classes is ignored. 
Conversely, existing multi-class land-cover benchmarks 
(\eg, LoveDA~\cite{wang2021loveda}, ISPRS Vaihingen/Potsdam~\cite{ISPRS_Vaihingen_2D, ISPRS_Potsdam_2D}) 
provide only raster masks, lacking vector geometry. 
No public dataset currently supports evaluating ACPV, defined in Sec.~\ref{sec:acpv}, as an all-class polygonal vectorization task with strict topological constraints, nor provides standardized \emph{global topology-consistency metrics} for such evaluation. 
As shown in Table \ref{tab:dataset_comparison}, the proposed \textbf{Deventer-512} fills this gap by providing 
high-resolution aerial imagery with topologically valid, multi-class vector annotations forming a single planar partition per patch, 
and a standardized evaluation protocol that measures both per-class polygonal quality and global topological consistency. 

\begin{table}[t]
  \centering
  \caption{
Comparison of representative datasets for \textbf{polygonal outline extraction} (top) and \textbf{land-cover mapping} (bottom). ``V'' indicates vector annotations; ``S'' shared-edge consistency; ``T'' global topology-consistency metrics; and ``C'' cadastral-aligned boundaries (as opposed to purely visual contours).
  }
  \vspace{-4pt}
  \label{tab:dataset_comparison}
  \renewcommand{\arraystretch}{0.9}
  \setlength{\tabcolsep}{10pt}  
  \resizebox{\columnwidth}{!}{
  \begin{tabular}{@{}l|cccccc@{}}
    \toprule
    Dataset & Multi-class & V & S & T & C \\
    \midrule
    Inria~\cite{maggiori2017can} & \ding{55} & \ding{55} & \ding{55} & \ding{55} & \checkmark \\
    WHU-Building~\cite{ji2018fully} & \ding{55} & \checkmark & \ding{55} & \ding{55} & \checkmark \\
    \midrule
    ISPRS Vaihingen~\cite{ISPRS_Vaihingen_2D} & \checkmark & \ding{55} & \ding{55} & \ding{55} & \ding{55} \\
    ISPRS Potsdam~\cite{ISPRS_Potsdam_2D} & \checkmark & \ding{55} & \ding{55} & \ding{55} & \ding{55} \\
    LoveDA~\cite{wang2021loveda} & \checkmark & \ding{55} & \ding{55} & \ding{55} & \ding{55} \\
    \midrule
    \textbf{Deventer-512 (Ours)} & \checkmark & \checkmark & \checkmark & \checkmark & \checkmark \\
    \bottomrule
  \end{tabular}
  }
  \vspace{-6pt}
\end{table}

\textbf{Deventer-512} is built from ortho-rectified aerial RGB imagery over the Deventer region in the eastern Netherlands, 
acquired from the national cadastral mapping service (\emph{Kadaster}) at a ground sampling distance of 0.3\,m. 
All annotated polygons within each $512{\times}512$ patch are organized to form a topology-consistent planar partition with shared boundaries (Sec.~\ref{sec:acpv}). 
Polygon annotations originate from official cadastral data and are refined through a topology-validation pipeline to ensure \emph{global topological consistency} (details in the supplementary material).

The dataset contains five land-cover classes, \ie \emph{buildings, roads, vegetation, water}, and \emph{unvegetated area}, each represented as a topologically valid polygon set. 
It comprises 1{,}716 training, 212 validation, and 220 testing patches, totaling 84{,}403 instances and 22{,}679 internal holes, with the most complex polygon containing up to 708 vertices. 
The polygons exhibit higher geometric complexity and vertex density than single-class building datasets, where object geometry is relatively simple. 
They are delineated following cadastral or administrative boundaries rather than purely visual ones, which introduces semantic–visual discrepancies, leading to semantically ambiguous boundaries that make the dataset particularly challenging. 
A more detailed description of the dataset is provided in the supplementary.

\subsection{Evaluation protocol and baselines}
\label{sec:eval_protocol}
\noindent\textbf{Evaluation Metrics.}
We evaluate polygonal quality under a unified metric suite spanning: \emph{semantic fidelity}, \emph{geometric accuracy}, \emph{vertex efficiency}, and \emph{topological consistency}. 
Details are provided in the supplementary material.

\vspace{2pt}
\noindent\textbf{Baselines.}
We evaluate five representative state-of-the-art polygonization methods over the three major paradigms discussed in Sec.~\ref{sec:related_work}:
\emph{DeepSnake}~\cite{peng2020deep} (a learned polygonization module based on active contour evolution),
\emph{FFL}~\cite{girard2021polygonal} (frame-field–guided polygonization),
\emph{TopDiG}~\cite{yang2023topdig} (transformer-based vertex detection and adjacency learning),
\emph{HiSup}~\cite{xu2023hisup} (hierarchical attraction-field supervision),
and \emph{GCP}~\cite{zhang2025global} (a global contour proposal network with transformer-based polygon refinement). 
They span learned polygon refinement, vertex extraction and adjacency learning, and geometric-field modeling, ensuring comprehensive coverage of existing polygonization paradigms. 
They are trained and evaluated on Deventer-512 using identical patch-based settings and evaluation metrics. 

\section{Experiments}
\label{sec:experiments}
\subsection{ACPV on Deventer-512}
We first evaluate the output vector basemap as a whole to verify compliance with the ACPV constraints in realizing seamless all-class vector map generation from aerial imagery. 
As shown in Table \ref{tab:deventer_topology_global}, when the single-class outputs of baselines are stitched together, they consistently exhibit non-zero inter- and intra-overlap rates, 
whereas ACPV-Net achieves zero measured gap and overlap rates and 100\% shared-edge consistency, confirming the seamless nature of the generated vector basemap. 
Detailed per-class overlap statistics are reported in the supplementary material. 

We further analyze per-class performance by extracting polygons of each land-cover class from this basemap and compare them with state-of-the-art single-class polygonization methods, each applied separately to individual land-cover classes. 
As shown in Table \ref{tab:deventer_512}, 
ACPV-Net produces all classes in one pass and achieves consistent improvement across all five classes in semantic fidelity, geometric accuracy, topological fidelity, and vertex efficiency, without any class-specific tuning, except for a slightly higher MTA on water (42.63 ours vs.~42.22 HiSup). 
This difference is likely due to our more compact vertex representation measured in N-ratio (0.92 ours vs.~3.26 HiSup) for representing irregular curved water body boundaries. 

\begin{figure}[t]
  \centering
  \subfloat[Input]{%
    \includegraphics[width=0.19\linewidth, trim={139mm 8mm 139mm 0mm},clip]{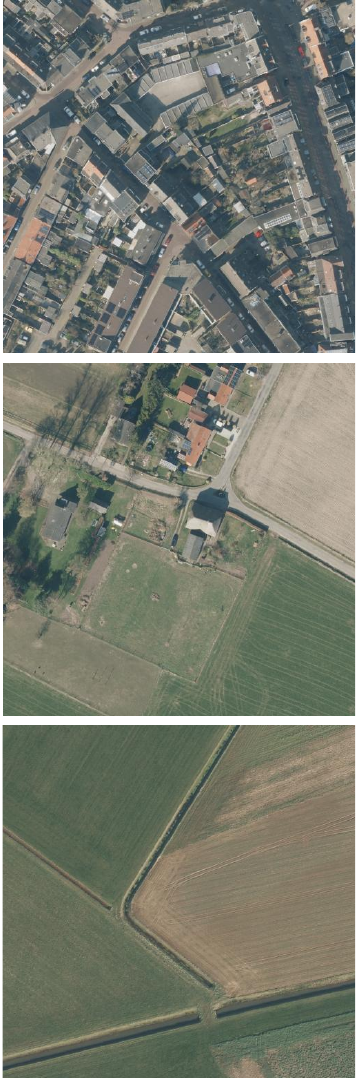}}\hfill
  \subfloat[GT]{%
    \includegraphics[width=0.19\linewidth, trim={139mm 8mm 139mm 0mm},clip]{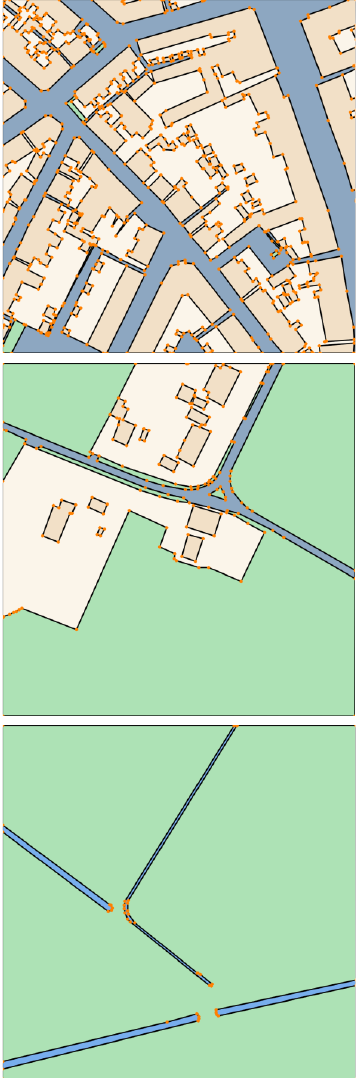}}\hfill
  \subfloat[TopDiG]{%
    \includegraphics[width=0.19\linewidth, trim={139mm 8mm 139mm 0mm},clip]{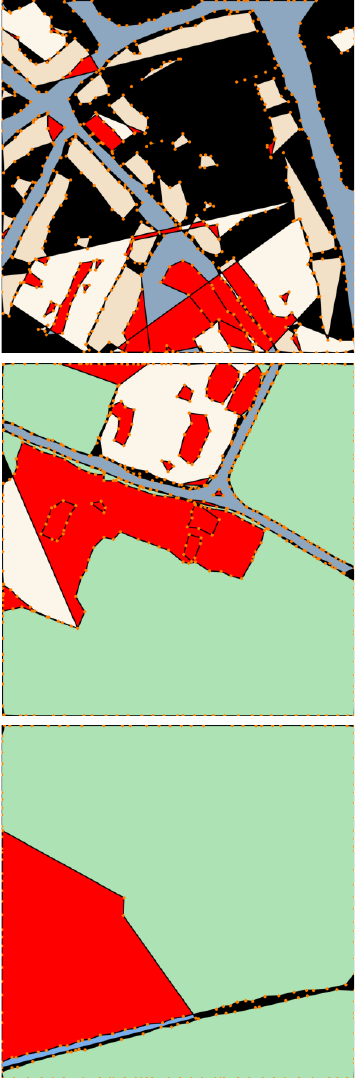}}\hfill
  \subfloat[HiSup]{%
    \includegraphics[width=0.19\linewidth, trim={139mm 8mm 139mm 0mm},clip]{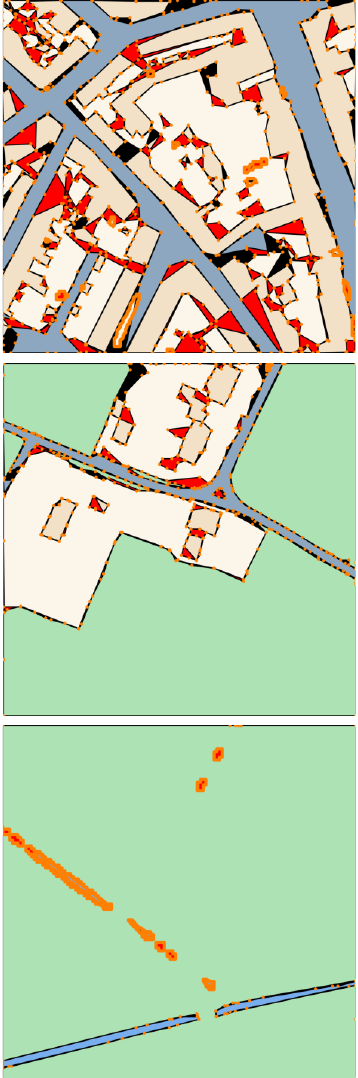}}\hfill
  \subfloat[Ours]{%
    \includegraphics[width=0.19\linewidth, trim={139mm 8mm 139mm 0mm},clip]{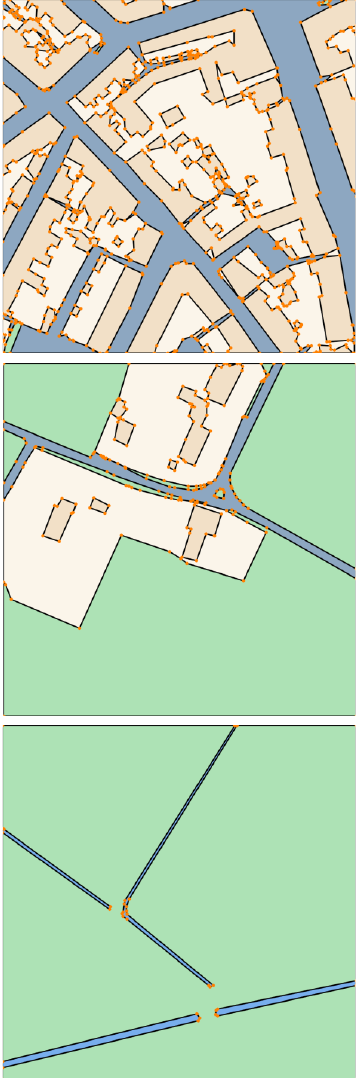}}\hfill
  \caption{
    Qualitative comparison on Deventer-512. 
The three rows show representative urban, suburban, and rural scenes, respectively. 
From left to right: aerial imagery, ground truth, TopDiG~\cite{yang2023topdig}, HiSup~\cite{xu2023hisup}, and Ours. 
Land-cover classes are color-coded; polygon outlines are drawn in black, vertices are highlighted with orange dots, 
and inter-class overlaps and gaps are marked in red and black, respectively.}
  \label{fig:deventer_512_visual_compare}
  \vspace{-6pt}
\end{figure}

Figure~\ref{fig:deventer_512_visual_compare} presents three representative scenes from the Deventer-512 test set: \emph{urban}, \emph{suburban}, and \emph{rural areas}. 
These examples also cover challenging visual conditions such as shadow, partial occlusion, and semantically ambiguous boundaries.
ACPV-Net predicts more accurate vector maps than the other models, illustrating the robustness of ACPV-Net under weak visual cues. 
Additional qualitative comparisons under more visual conditions and baselines are provided in the supplementary material. 

\begin{table}[t]
  \centering
  \caption{
  \textbf{Global topological consistency on Deventer-512.} 
  Metrics include gap rate (Gap), inter-class overlap rate (Inter), intra-class overlap rate (Intra) and shared-edge consistency rate (Shared). 
  Best and second-best values are highlighted in \textbf{bold} and \underline{underlined}, respectively. 
  All percentage values are reported to two decimal places.
  }
  \label{tab:deventer_topology_global}
  \renewcommand{\arraystretch}{0.9}
  \setlength{\tabcolsep}{8pt}  
  \resizebox{\columnwidth}{!}{%
  \begin{tabular}{l|cccc}
    \toprule
    Method & Gap $\downarrow$ & Inter $\downarrow$ & Intra $\downarrow$ & Shared $\uparrow$ \\
    \midrule
    DeepSnake \cite{peng2020deep}\conf{CVPR}{20} & 12.41 & 68.86 & 51.16 & \underline{38.73} \\
    FFL \cite{girard2021polygonal}\conf{CVPR}{21} & 5.48 & 29.17 & \underline{0.08} & 9.20 \\
    TopDiG \cite{yang2023topdig}\conf{CVPR}{23} & 8.47 & 13.57 & \textbf{0.00} & 10.81 \\
    HiSup \cite{xu2023hisup}\conf{ISPRS}{23} & \underline{5.43} & \underline{4.50} & \textbf{0.00} & 25.73 \\
    GCP \cite{zhang2025global}\conf{TGRS}{25} & 8.75 & 10.39 & 41.91 & 20.25 \\
    \midrule
    \cellcolor{oursbg} ACPV-Net (Ours) & \cellcolor{oursbg}\textbf{0.00} & \cellcolor{oursbg}\textbf{0.00} & \cellcolor{oursbg}\textbf{0.00} & \cellcolor{oursbg}\textbf{100.00} \\
    \bottomrule
  \end{tabular}%
  }
\end{table}

\begin{table*}[t]
  \caption{\textbf{Quantitative comparison on Deventer-512.} Best and second-best values are highlighted in \textbf{bold} and \underline{underlined}, respectively. APLS is evaluated only on elongated classes (roads and water bodies).}
  \vspace{-4pt}
  \setlength{\tabcolsep}{12pt}  
  \label{tab:deventer_512}
  \centering
  \renewcommand{\arraystretch}{0.85}
  \resizebox{\textwidth}{!}{%
  \begin{tabular}{l| l| cc| cc| cc| ccc}         
    \toprule
    \multirow{2}{*}{Class} & \multirow{2}{*}{Method} &
    \multicolumn{2}{c|}{Semantic fidelity} &
    \multicolumn{2}{c|}{Vertex efficiency} &
    \multicolumn{2}{c|}{Geometric accuracy} &
    \multicolumn{3}{c}{Topological fidelity} \\    
    \cmidrule(lr){3-4} \cmidrule(lr){5-6} \cmidrule(lr){7-8} \cmidrule(lr){9-11}   
     & & IoU$\uparrow$ & B-IoU$\uparrow$ & N-ratio$\rightarrow$1 & C-IoU$\uparrow$ & PoLiS$\downarrow$ & MTA$\downarrow$ & APLS$\uparrow$ & $\chi$-Err$\downarrow$ & $\beta$-Err$\downarrow$  \\
    \midrule
    \multirow{6}{*}{Road}
      & DeepSnake \cite{peng2020deep} \conf{CVPR}{20} & 10.35 & 7.90 & 74.91 & 4.58 & 33.18 & 53.38 & 64.85 & 7.82 & 10.11 \\
      & FFL \cite{girard2021polygonal} \conf{CVPR}{21} & 62.00 & 58.08 & 3.74 & 42.28 & 11.48 & \underline{44.05} & 78.53 & 15.31 & 15.40 \\
      & TopDiG \cite{yang2023topdig} \conf{CVPR}{23} & 58.75 & 55.95 & 1.80 & 44.06 & 10.28 & 44.27 & 70.65 & 5.90 & 6.54 \\
      & HiSup \cite{xu2023hisup} \conf{ISPRS}{23} & \underline{73.92} & \underline{69.29} & 2.00 & \underline{57.36} & \underline{4.97} & 44.84 & \underline{81.06} & 5.16 & 5.45 \\
      & GCP \cite{zhang2025global} \conf{TGRS}{25} & 59.93 & 56.58 & \underline{0.86} & 44.96 & 9.18 & 49.19 & 73.00 & \underline{4.38} & \underline{5.18} \\
      \cmidrule(lr){2-11}
      & \cellcolor{oursbg}Ours & \cellcolor{oursbg}\textbf{76.01} & \cellcolor{oursbg}\textbf{73.32} & \cellcolor{oursbg}\textbf{1.07} & \cellcolor{oursbg}\textbf{68.22} & \cellcolor{oursbg}\textbf{4.44} & \cellcolor{oursbg}\textbf{43.85} & \cellcolor{oursbg}\textbf{83.52} & \cellcolor{oursbg}\textbf{3.68} & \cellcolor{oursbg}\textbf{4.04} \\
    \midrule
    \multirow{6}{*}{Building}
      & DeepSnake \cite{peng2020deep} \conf{CVPR}{20} & 34.24 & 29.96 & 30.00 & 9.64 & 4.41 & 50.38 & --- & 7.39 & 8.45 \\
      & FFL \cite{girard2021polygonal} \conf{CVPR}{21} & 63.69 & 60.91 & 8.69 & 40.56 & 3.21 & 48.12 & --- & 39.71 & 42.60 \\
      & TopDiG \cite{yang2023topdig} \conf{CVPR}{23} & 72.56 & 69.03 & \underline{1.07} & 63.12 & 3.37 & 45.97 & --- & 7.21 & 8.37 \\
      & HiSup \cite{xu2023hisup} \conf{ISPRS}{23} & \underline{81.22} & \underline{78.06} & 1.55 & \underline{70.60} & \underline{2.23} & \underline{42.59} & --- & \underline{6.82} & 7.70 \\
      & GCP \cite{zhang2025global} \conf{TGRS}{25} & 73.97 & 70.41 & 1.79 & 58.59 & 2.91 & 47.37 & --- & 6.84 & \underline{7.45} \\
      \cmidrule(lr){2-11}
      & \cellcolor{oursbg}Ours & \cellcolor{oursbg}\textbf{82.08} & \cellcolor{oursbg}\textbf{79.50} & \cellcolor{oursbg}\textbf{1.00} & \cellcolor{oursbg}\textbf{77.24} & \cellcolor{oursbg}\textbf{1.76} & \cellcolor{oursbg}\textbf{39.39} & \cellcolor{oursbg}--- & \cellcolor{oursbg}\textbf{2.32} & \cellcolor{oursbg}\textbf{2.72} \\
    \midrule
    \multirow{6}{*}{Unvegetated}
      & DeepSnake \cite{peng2020deep} \conf{CVPR}{20} & 21.06 & 10.84 & 68.93 & 3.31 & 17.32 & 53.90 & --- & 8.25 & 11.85 \\
      & FFL \cite{girard2021polygonal} \conf{CVPR}{21} & 43.91 & 35.56 & 30.62 & 24.99 & 19.27 & 47.40 & --- & 84.35 & 91.59 \\
      & TopDiG \cite{yang2023topdig} \conf{CVPR}{23} & 49.26 & 39.92 & 0.77 & 40.61 & 12.73 & 48.44 & --- & 6.30 & 10.28 \\
      & HiSup \cite{xu2023hisup} \conf{ISPRS}{23} & \underline{61.46} & \underline{55.37} & 1.93 & \underline{51.79} & \underline{8.66} & \underline{47.17} & --- & \underline{4.22} & \underline{5.34} \\
      & GCP \cite{zhang2025global} \conf{TGRS}{25} & 47.85 & 38.39 & \underline{1.23} & 41.09 & 10.71 & 53.67 & --- & 6.10 & 8.09 \\
      \cmidrule(lr){2-11}
      & \cellcolor{oursbg}Ours & \cellcolor{oursbg}\textbf{66.89} & \cellcolor{oursbg}\textbf{61.51} & \cellcolor{oursbg}\textbf{1.09} & \cellcolor{oursbg}\textbf{61.79} & \cellcolor{oursbg}\textbf{6.41} & \cellcolor{oursbg}\textbf{46.79} & \cellcolor{oursbg}--- & \cellcolor{oursbg}\textbf{2.80} & \cellcolor{oursbg}\textbf{3.62} \\
    \midrule
    \multirow{6}{*}{Vegetation}
      & DeepSnake \cite{peng2020deep} \conf{CVPR}{20} & 51.18 & 25.88 & 58.81 & 3.70 & 16.63 & 51.89 & --- & 16.89 & 17.96 \\
      & FFL \cite{girard2021polygonal} \conf{CVPR}{21} & 61.19 & 39.18 & 4.12 & 29.23 & 15.46 & 46.35 & --- & 33.65 & 33.95 \\
      & TopDiG \cite{yang2023topdig} \conf{CVPR}{23} & 70.02 & 49.64 & 2.67 & 44.09 & 11.35 & 45.79 & --- & 9.31 & 11.43 \\
      & HiSup \cite{xu2023hisup} \conf{ISPRS}{23} & \underline{75.70} & \underline{57.85} & 3.54 & \underline{46.07} & \underline{7.51} & \underline{41.86} & --- & \underline{8.53} & \underline{9.70} \\
      & GCP \cite{zhang2025global} \conf{TGRS}{25} & 70.37 & 49.65 & \underline{2.57} & 45.24 & 9.00 & 49.29 & --- & 12.97 & 14.05 \\
      \cmidrule(lr){2-11}
      & \cellcolor{oursbg}Ours & \cellcolor{oursbg}\textbf{80.05} & \cellcolor{oursbg}\textbf{64.50} & \cellcolor{oursbg}\textbf{0.98} & \cellcolor{oursbg}\textbf{67.55} & \cellcolor{oursbg}\textbf{6.22} & \cellcolor{oursbg}\textbf{41.43} & \cellcolor{oursbg}\textbf{---} & \cellcolor{oursbg}\textbf{6.62} & \cellcolor{oursbg}\textbf{7.53} \\
    \midrule
    \multirow{6}{*}{Water}
      & DeepSnake \cite{peng2020deep} \conf{CVPR}{20} & 27.27 & 20.93 & 116.05 & 14.87 & 14.62 & 53.55 & 56.05 & 5.57 & 6.83 \\
      & FFL \cite{girard2021polygonal} \conf{CVPR}{21} & 54.73 & 47.02 & 5.53 & 41.72 & 7.92 & 44.91 & \underline{61.22} & 11.24 & 11.29 \\
      & TopDiG \cite{yang2023topdig} \conf{CVPR}{23} & 43.23 & 39.65 & 0.36 & 33.61 & 9.14 & 43.58 & 30.75 & 1.93 & 2.05 \\
      & HiSup \cite{xu2023hisup} \conf{ISPRS}{23} & \underline{64.84} & \underline{58.23} & 3.26 & \underline{51.79} & \underline{4.29} & \textbf{42.22} & 58.40 & 1.77 & 1.86 \\
      & GCP \cite{zhang2025global} \conf{TGRS}{25} & 52.74 & 45.87 & \textbf{1.08} & 44.52 & 5.00 & 46.24 & 60.85 & \underline{1.55} & \underline{1.64} \\
      \cmidrule(lr){2-11}
      & \cellcolor{oursbg}Ours & \cellcolor{oursbg}\textbf{67.96} & \cellcolor{oursbg}\textbf{62.05} & \cellcolor{oursbg}\textbf{0.92} & \cellcolor{oursbg}\textbf{59.53} & \cellcolor{oursbg}\textbf{3.83} & \cellcolor{oursbg}\underline{42.63} & \cellcolor{oursbg}\textbf{68.81} & \cellcolor{oursbg}{\textbf{1.28}} & \cellcolor{oursbg}\textbf{1.34} \\
    \bottomrule
  \end{tabular}%
  }
\end{table*}

\subsection{Single-class Polygonal Vectorization}
\begin{table}[ht]
  \caption{Quantitative comparison on WHU-Building. Best values are in \textbf{bold} and second-best values are \underline{underlined}.}
  \vspace{-4pt}
  \label{tab:whu_building}
  \centering
  \setlength{\tabcolsep}{2pt}
  \resizebox{\columnwidth}{!}{
  \begin{tabular}{l| cc| cc| cc| cc}
    \toprule
    \multirow{2}{*}{Method} &
    \multicolumn{2}{c|}{Sem. fidelity} &
    \multicolumn{2}{c|}{Vertex efficiency} &
    \multicolumn{2}{c|}{Geo. accuracy} &
    \multicolumn{2}{c}{Topo. fidelity} \\
    \cmidrule(lr){2-3} \cmidrule(lr){4-5} \cmidrule(lr){6-7} \cmidrule(lr){8-9}
     & IoU$\uparrow$ & B-IoU$\uparrow$ & N-ratio$\rightarrow$1 & C-IoU$\uparrow$ & PoLiS$\downarrow$ & MTA$\downarrow$ & $\chi$-Err$\downarrow$ & $\beta$-Err$\downarrow$  \\
    \midrule
      DeepSnake \cite{peng2020deep} & 71.39 & 64.10 & 2.05 & 54.21 & 2.17 & 46.85 & 4.43 & 6.29  \\
      FFL \cite{girard2021polygonal} & 80.86 & 74.63 & 3.68 & 46.09 & 2.13 & 38.86 & 12.33 & 14.88  \\
      TopDiG \cite{yang2023topdig} & 83.79 & 77.73 & 1.95 & 62.25 & 1.99 & 38.82 & 2.13 & 2.56  \\
      HiSup \cite{xu2023hisup} & \underline{87.63} & \underline{82.82} & \underline{1.93} & \underline{67.15} &  \underline{1.40} & \underline{35.27} & \underline{1.95} & \underline{2.28}  \\
      GCP \cite{zhang2025global} & 83.28 & 77.58 & 2.11 & 60.51 & 1.66 & 34.96 & 3.20 & 4.39  \\
      \midrule
      \cellcolor{oursbg}Ours & \cellcolor{oursbg}\textbf{88.50} & \cellcolor{oursbg}\textbf{83.39} & \cellcolor{oursbg}\textbf{1.07} & \cellcolor{oursbg}\textbf{81.45} & \cellcolor{oursbg}\textbf{1.38} & \cellcolor{oursbg}\textbf{34.85} & \cellcolor{oursbg}\textbf{1.60} & \cellcolor{oursbg}\textbf{1.81} \\
    \bottomrule
  \end{tabular}
  }
  \vspace{-6pt}
\end{table}

To further validate the applicability of ACPV-Net beyond multi-class mapping, 
we directly apply the same architecture to single-class polygonal vectorization without any structural modification. 
We select the WHU-Building dataset \cite{ji2018fully} for this experiment because \textit{building} is the most common and structurally representative category in aerial imagery, 
and the dataset provides large-scale, high-resolution images with reliable annotations. 
As summarized in Table~\ref{tab:whu_building}, ACPV-Net achieves the best-reported results on WHU-Building against building-specific single-class polygonization baselines, demonstrating that ACPV-Net is not limited to seamless multi-class vector map generation.
It also serves as a unified and flexible framework for single-class polygonal outline extraction. 
Representative visual results are provided in the supplementary material. 

It is worth mentioning that, to examine cross-region robustness, we apply the same model trained on Deventer-512 to high-resolution aerial imagery from cities in other countries and sensors without fine-tuning. 
The results (see supplementary) show consistent topological validity and reasonable polygon regularity in visually similar regions.

\subsection{Ablation Studies}
\label{sec:ablation}
We conduct comprehensive ablations to verify the effectiveness of each key design component in ACPV-Net. 
Specifically, we analyze 
(i) how semantically supervised conditioning (SSC) suppresses artifacts and enhances semantic–geometric alignment, 
(ii) how distributional vertex modeling via latent reconstruction improves robustness and vertex precision under weak/ambiguous visual cues, 
and (iii) how the deterministic reconstruction ensures topological consistency by design. 
All experiments are performed on Deventer-512 under identical settings (see implementation details in the supplementary).

\paragraph{SSC and Distributional Vertex Modeling.}

\begin{figure}[t!]
  \centering
  \includegraphics[width=\linewidth,trim={94mm 48mm 94mm 48mm},clip]{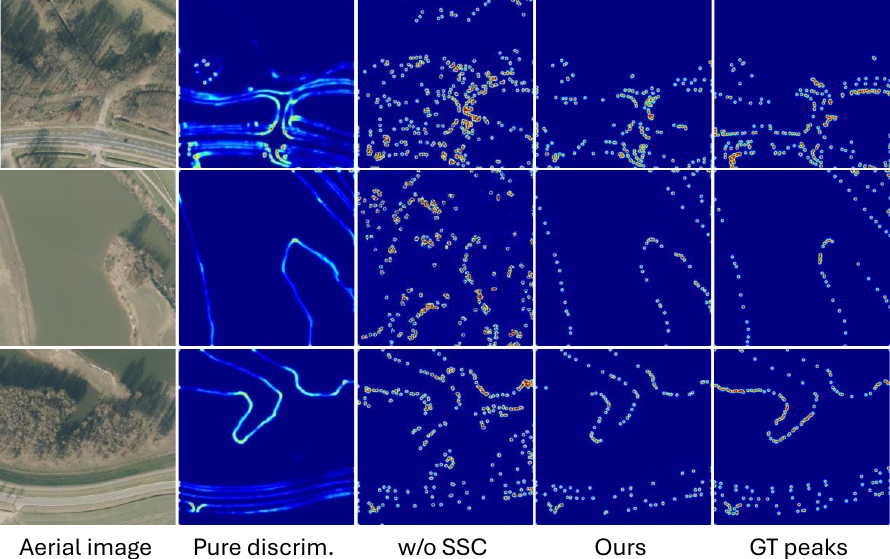}
  \caption{Vertex activations under weak/ambiguous visual cues and along smooth boundaries (cartographic convention cases). From left to right: aerial image, pure discriminative decoding, without semantic supervision (No-SSC), ours, and ground truth. 
  }
  \label{fig:weak_visual_cues}
  \vspace{-8pt}
\end{figure}

We conduct two controlled experiments to isolate the effects of \textbf{(i)} explicit semantic supervision (SSC) and \textbf{(ii)} distributional vertex modeling through latent reconstruction.

\noindent\textbf{(i) No explicit semantic supervision (No-SSC).} 
We disable the gradient flow of the segmentation loss $\mathcal{L}_{\text{seg}}$ to the conditioning encoder $S_\psi$, 
so that its features are optimized only through the diffusion objective. 
This setting reduces the model to a typical conditional diffusion setup (\eg \cite{rombach2022high, zhang2023adding, yang2024pixel, zhao2025local}) without explicit task-specific supervision. 
The segmentation head remains active, producing $\hat{M}$ for fair comparison. 
Without explicit semantic guidance, the model produces numerous false-positive vertex activations, many of which appear inside homogeneous regions rather than along class-consistent boundaries 
(see Fig.~\ref{fig:weak_visual_cues}). 
We quantify this misalignment using the \emph{Vertex–Boundary Alignment (V2B\@$\delta$)} metric, the fraction of predicted vertices that fall within $\tau$ pixels of their class-consistent boundaries. 
As shown in Table~\ref{tab:ssc_v2b}, removing $\mathcal{L}_{\text{seg}}$ leads to a clear drop in alignment, confirming that explicit semantic supervision enhances semantic–geometric consistency.

\begin{table}[t]
\centering
\caption{Vertex–Boundary Alignment (V2B@$\delta$) comparison between \emph{No-SSC} and \emph{Full SSC}.
Global boundaries are computed as multi-class label transitions from the raster mask. 
\textbf{Avg(2,4,6)} denotes the mean of V2B@$2$, @$4$, and @$6$. 
}
\vspace{-4pt}
\label{tab:ssc_v2b}
\renewcommand{\arraystretch}{0.9}
\setlength{\tabcolsep}{10pt}  
\resizebox{\columnwidth}{!}{
\begin{tabular}{@{}l|cccc@{}}
\toprule
Variant & V2B@2$\uparrow$ & V2B@4$\uparrow$ & V2B@6$\uparrow$ & Avg@{2,4,6}$\uparrow$ \\
\midrule
No-SSC & 0.38 & 0.48 & 0.53 & 0.46 \\
\textbf{Full SSC (ours)} & \textbf{0.78} & \textbf{0.87} & \textbf{0.90} & \textbf{0.85} \\
\bottomrule
\end{tabular}
}
\end{table}

\noindent\textbf{(ii) Pure discriminative decoding.}
To assess the impact of distributional vertex modeling, we replace the latent reconstruction with a direct pixel-space decoder (\emph{pure discriminative}) adopting the ViTPose~\cite{xu2022vitpose} design as a strong baseline. 
We analyze vertex peak shapes using three intuitive metrics--\emph{FWHM} (peak width), \emph{Area@0.5} (size of high-response region), and \emph{Sharpness} (local contrast)--with full computation details provided in the supplementary material. 
As shown in Fig.~\ref{fig:violin_peakshape} and Table~\ref{tab:peakshape_stats}, the diffusion-based reconstruction yields sharper, more isotropic, and spatially compact peaks (smaller FWHM, lower high-response area, and higher sharpness), whereas the discriminative head produces broad or band-like responses (see Fig.~\ref{fig:weak_visual_cues}). 
These results demonstrate that our approach effectively handles the challenges of weak visual cues and smooth boundaries relying on cartographic conventions highlighted in Sec.~\ref{sec:intro}. 
\begin{figure}[t]
\centering
\subfloat[$\text{FWHM}_x$]{%
\includegraphics[width=0.48\columnwidth,trim={32mm 0mm 32mm 0mm},clip]{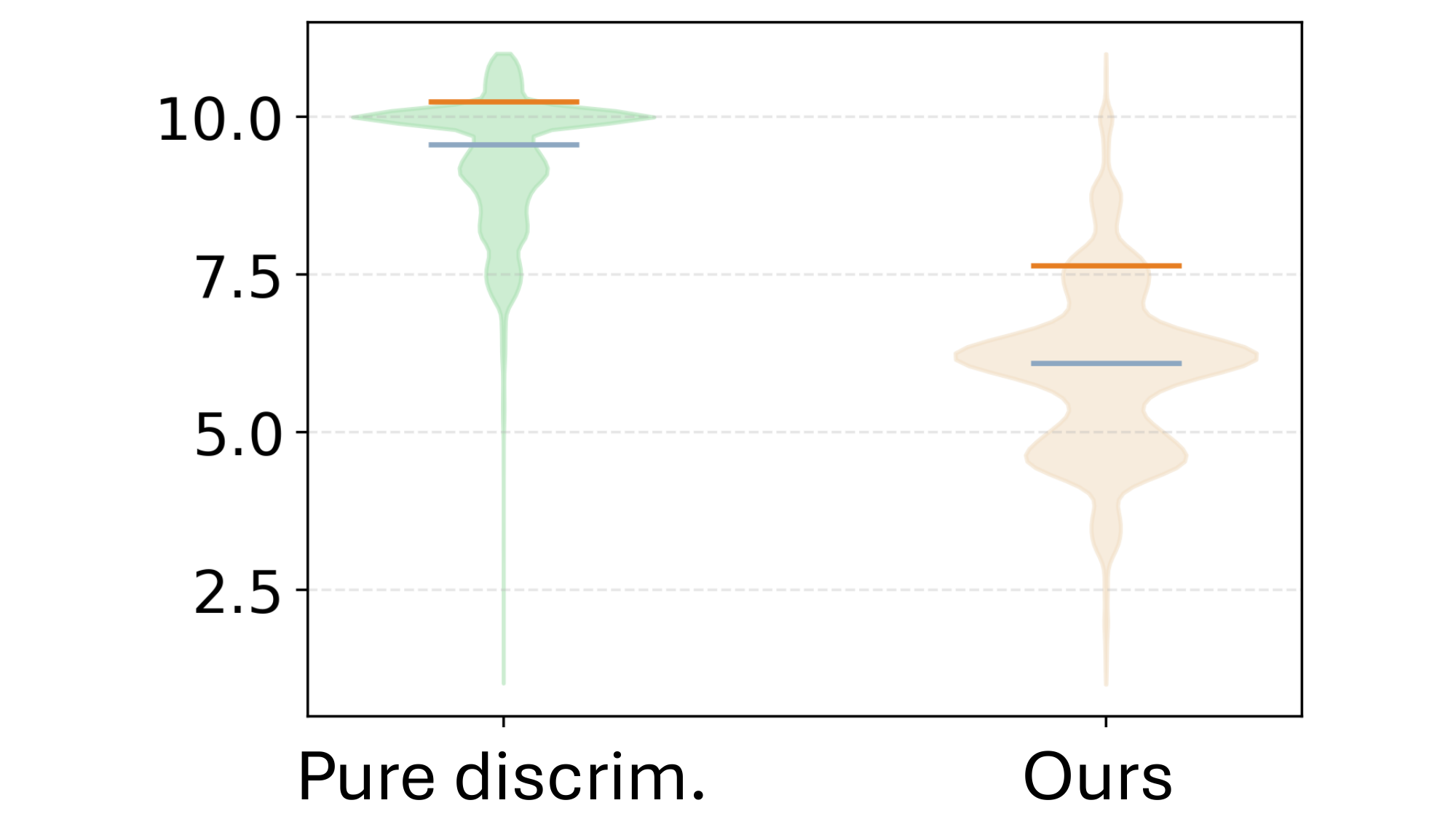}}\hfill
\subfloat[$\text{FWHM}_y$]{%
\includegraphics[width=0.48\columnwidth,trim={32mm 0mm 32mm 0mm},clip]{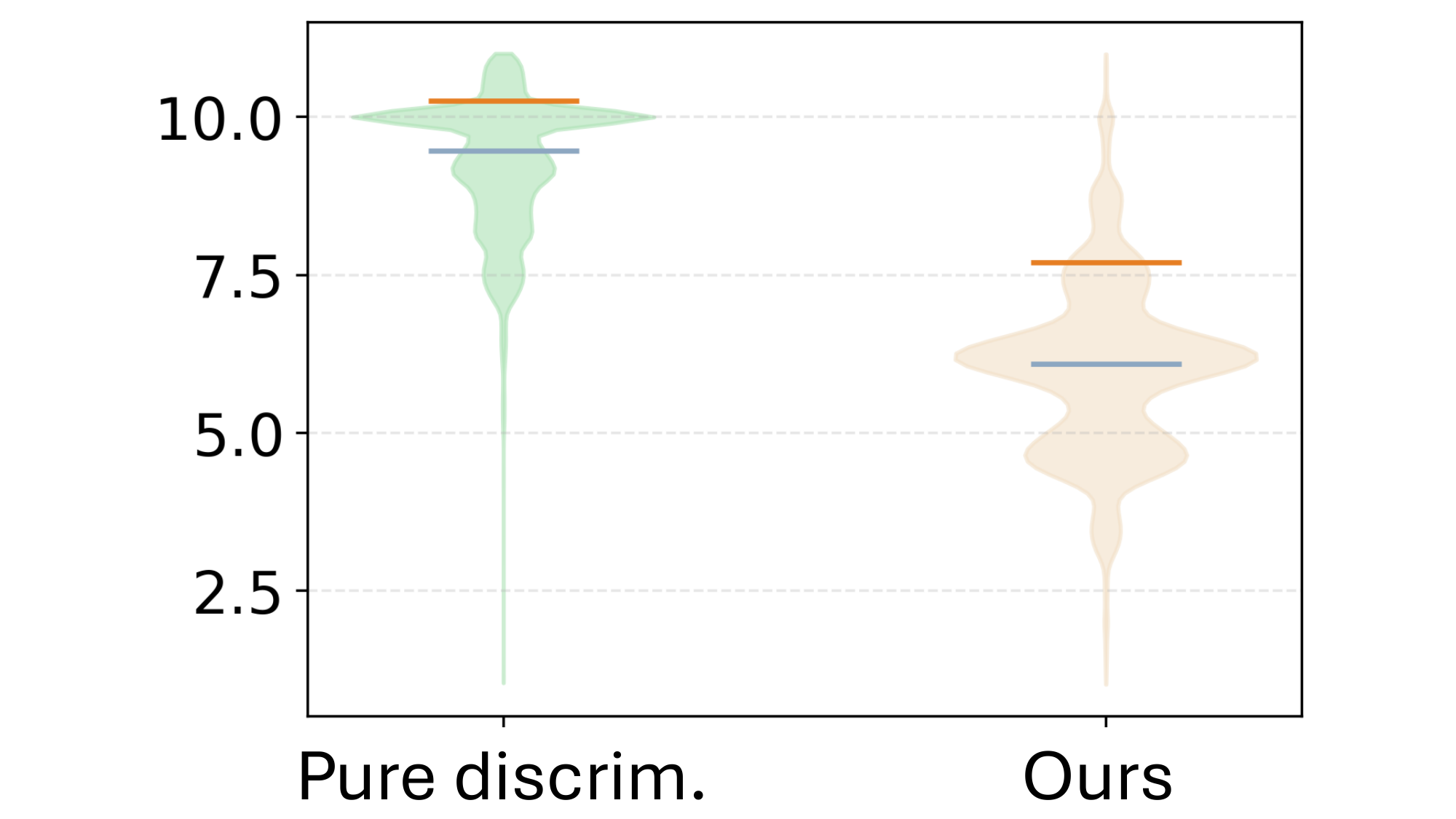}}
\par\smallskip
\subfloat[Area@0.5]{%
\includegraphics[width=0.48\columnwidth,trim={32mm 0mm 32mm 0mm},clip]{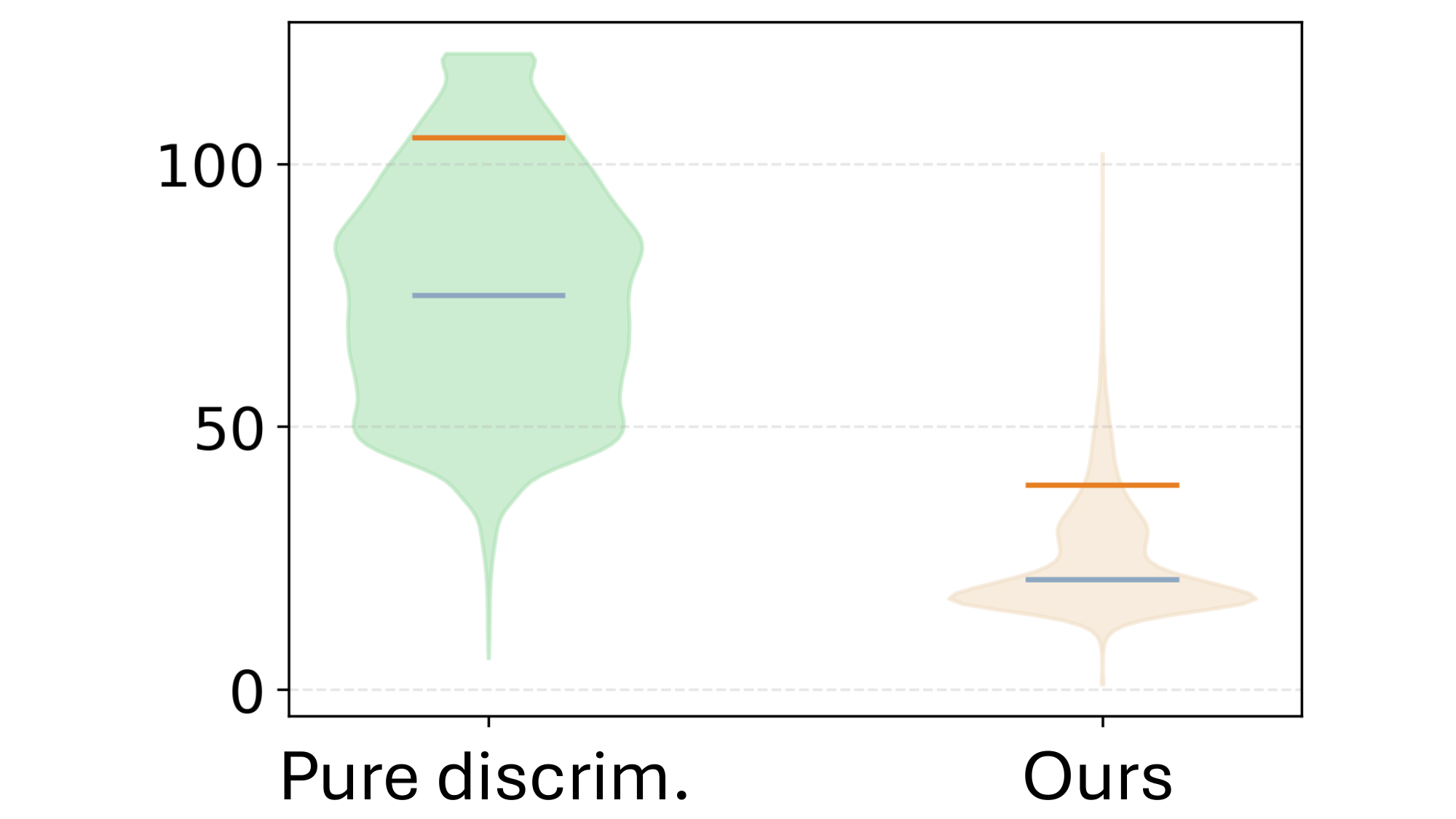}}\hfill
\subfloat[Sharpness]{%
\includegraphics[width=0.48\columnwidth,trim={32mm 0mm 32mm 0mm},clip]{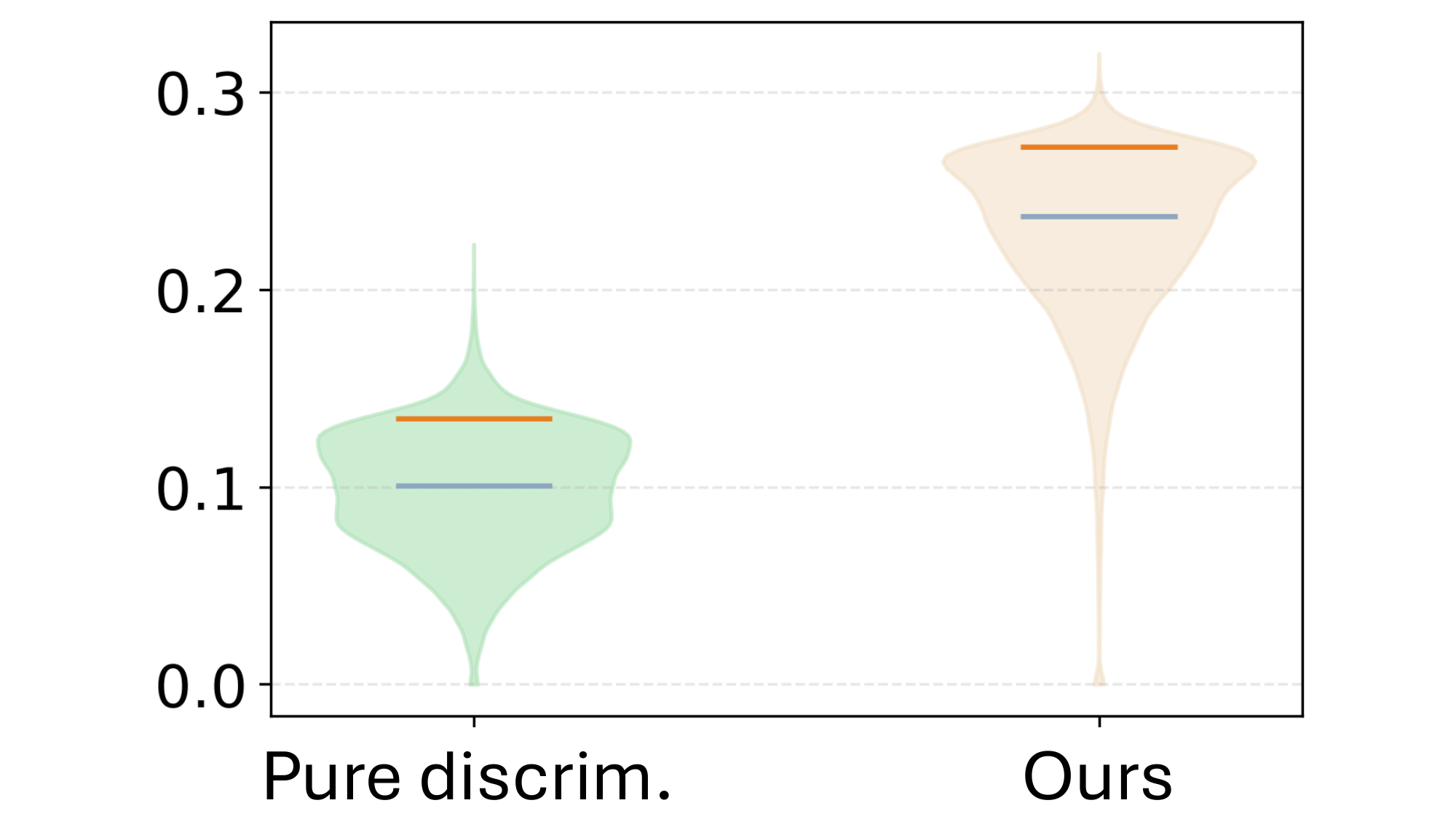}}
\caption{Vertex peak-shape comparison between the \emph{pure discriminative} baseline and our \emph{distributional reconstruction}.
Lower values of FWHM and Area@0.5 indicate sharper and more compact peaks, while higher Sharpness reflects stronger local contrast. The blue/red bar denotes the median/90th percentile.}
\label{fig:violin_peakshape}
\end{figure}

\begin{table}[t]
\centering
\caption{Summary statistics (median and 90th percentile) of vertex peak-shape metrics.}
\vspace{-4pt}
\label{tab:peakshape_stats}
\renewcommand{\arraystretch}{0.9}
\setlength{\tabcolsep}{13pt}
\footnotesize
\begin{tabular}{@{}l|cc|cc@{}}
\toprule
Metric & \multicolumn{2}{c|}{Pure discrim.} & \multicolumn{2}{c}{Ours (dist. model)} \\
\cmidrule(lr){2-3} \cmidrule(l){4-5}
& Median & $P_{90}$ & Median & $P_{90}$ \\
\midrule
FWHM$_x$ $\downarrow$ & 9.55 & 10.23 & \textbf{6.09} & \textbf{7.64} \\
FWHM$_y$ $\downarrow$ & 9.46 & 10.25 & \textbf{6.09} & \textbf{7.70} \\
Area@0.5 $\downarrow$ & 75.0 & 105.0 & \textbf{21.0} & \textbf{39.0} \\
Sharpness $\uparrow$ & 0.10 & 0.14 & \textbf{0.24} & \textbf{0.27} \\
\bottomrule
\end{tabular}%
\vspace{-6pt}
\end{table}

We also evaluate robustness under varying NMS thresholds. 
The distributional model shows minimal variance across all polygonal metrics, 
whereas the discriminative baseline fluctuates notably (detailed in the supplementary).

\vspace{2pt}
\noindent\textbf{Ablation on Topological Reconstruction.}
\label{sec:ablation_topo_recon}
This evaluates the effectiveness of 
\textbf{(i)} the planar structural precondition established by the \emph{overdense PSLG construction}, 
and \textbf{(ii)} the geometric simplification enabled by the \emph{vertex-guided subset selection} (VSS). 
All variants share the same segmentation and vertex priors $(\hat{M},\hat{Y})$ for fair comparison.

\begin{table}[t]
  \centering
  \caption{Ablation of Topology-consistency on the reconstruction stage. Metrics are averaged over DP tolerances $\varepsilon\!\in\!\{1,2,3,4\}$.}
  \vspace{-8pt}
  \renewcommand{\arraystretch}{0.85}
  \label{tab:topo_consistency_ablation}
  \resizebox{\columnwidth}{!}{
  \begin{tabular}{@{}l|cccc@{}}
    \toprule
    Method & Gap$\downarrow$ & Inter$\downarrow$ & Intra$\downarrow$ & Shared$\uparrow$ \\
    \midrule
    No PSLG + DP (w/o VSS) & 1.04 & 0.14 & 0.00 & 15.38 \\
    PSLG + DP (w/o VSS) & \textbf{0.00} & \textbf{0.00} & \textbf{0.00} & \textbf{100.00} \\
    \bottomrule
  \end{tabular}}
  \vspace{-8pt}
\end{table}

\noindent\textbf{(i) Importance of the planar structural precondition.}
We compare the full PSLG-based reconstruction (``PSLG + DP, without VSS'') against a variant that omits the PSLG and polygonizes each semantic class independently from its mask boundaries 
(``No PSLG + DP, without VSS''). 
The goal is to validate the necessity of building a globally planar straight-line graph as required by Proposition~\ref{prop:acpv-sufficient}. 
As shown in Table~\ref{tab:topo_consistency_ablation}, removing the PSLG destroys the planar structural precondition, leading to severe violations in shared edge consistency. 
In contrast, the PSLG enforces a single global planar graph on which faces are labeled, serving as the essential precondition for realizing Proposition~\ref{prop:acpv-sufficient} and ensuring topology consistency \emph{by design}.

\noindent\textbf{(ii) Effect of VSS.}
We also compare our vertex-guided subset selection (VSS) with the traditional geometry-only simplification algorithm Douglas–Peucker (DP) \cite{douglas1973algorithms}. 
While DP minimizes geometric deviation, it is highly sensitive to the tolerance parameter and lacks data-driven saliency. 
Our VSS leverages learned vertices for adaptive keypoint preservation and yields lower redundancy at comparable or lower geometric error (see supplementary for details).

\section{Conclusion}
We presented \textbf{ACPV-Net}, the first fully automatic framework for seamless and topologically consistent vector basemap generation from aerial imagery. 
Departing from existing single-class polygonal outline extraction methods, our approach performs all-class polygonal vectorization in a single run. 
On the new \textbf{Deventer-512} benchmark, ACPV-Net achieves gap-/overlap-free topology with shared-edge consistency, and consistently outperforms strong single-class baselines across all land-cover classes. 
It also generalizes to single-class polygonal extraction, achieving the best results on the WHU-Building dataset. 
We believe this work represents an important step toward fully automated, large-scale, and topologically consistent vector map generation in remote sensing.

\section{Acknowledgment}
\label{sec:acknowledgment}

This publication is part of the project “Learning from old maps to create new ones”, with project number 19206 of the Open Technology Programme, which is financed by the Dutch Research Council (NWO), The Netherlands. We thank the members of the project for their support in data preparation.

{
    \small
    \bibliographystyle{ieeenat_fullname}
    \bibliography{main}
}


\end{document}


\maketitlesupplementary


\renewcommand{\thefigure}{S\arabic{figure}}
\renewcommand{\thetable}{S\arabic{table}}
\renewcommand{\theequation}{S\arabic{equation}}
\setcounter{figure}{0}\setcounter{table}{0}\setcounter{equation}{0}





\section*{Overview}
In this Appendix, we provide the following information:
Sec.~\ref{sec:sup_dataset} provides more details of the Deventer-512 dataset; 
Sec.~\ref{sec:eval} specifies the evaluation metrics, including per-class polygon quality metrics, global topology-consistency metrics, and vertex–boundary alignment and peak-shape metrics;
Sec.~\ref{sec:sup_method} provides additional details of the methodology, including the detailed architecture of SSC, the construction and simplification of the PSLG, and the proof of the sufficient condition for the topological reconstruction;
Sec.~\ref{sec:sup_impl} details the implementation and training, including data preprocessing, ACPV-Net and baseline training for fair comparison; 
Sec.~\ref{sec:sup_extended} includes more ablation studies and extended experimental results, as well as additional evaluations on the Shanghai dataset and TopDiG metrics on Deventer-512;
Lastly, Sec.~\ref{sec:sup_more_figs} closes this supplementary with additional qualitative visualizations and discussion with limitations.

\section{Dataset Details for Deventer-512}
\label{sec:sup_dataset}

\subsection{Vector Annotation and Topology Validation}
The raw polygon annotations of Deventer-512 originate from the official Dutch topographic map, the \textit{Basisregistratie Grootschalige Topografie} (BGT) \cite{Kadaster2022Dataset:BGT}, which provides polygon-based representations of multiple land-cover and land-use categories, and supports mapping at scales between 1:500 and 1:5000. Its standard positional accuracy is reported as 20\,cm for objects with a high absolute positional accuracy, and 40\,cm for objects with a low absolute positional accuracy. 
The BGT is created and maintained by various governmental bodies, with revision intervals typically ranging from 6 to 18 months \cite{GRIFT202491}.

Because the outlines of different land-cover classes are delineated independently, small geometric inconsistencies may occur along shared boundaries, occasionally producing microscopic gaps or overlaps.
To improve topological validity at the patch level, we apply a lightweight topology-validation procedure.  
Microscopic gaps and overlaps are detected using standard vector overlay operations (\eg, union, intersection), after which adjacent vertices within a tiny geometric tolerance are merged into a single endpoint. 
This improves shared-edge consistency while preserving the semantic regions at meaningful geometric scales. 
The resulting cleaned polygon network forms a coherent planar straight-line graph representing a topology-consistent planar partition with shared boundaries. 
These polygons serve as the ground-truth annotations in all experiments.

\subsection{Geometric Statistics and Complexity}
\label{sec:sup_dataset_stats}

\noindent\textbf{Per-class Geometric Statistics.}
Table~\ref{tab:stats_poly_complexity} reports per-class polygon statistics on the training set, including instance counts, vertex-count distribution (median, $P_{90}$, max), hole counts, and vertex density. 
The maximum vertex count ranges from 176 (buildings) to 708 (roads), indicating a wide span of geometric complexity with a strongly long-tailed distribution, particularly for roads and unvegetated regions. 
Hole counts also differ by orders of magnitude, reflecting highly non-uniform structural complexity across classes.

We additionally compute vertex density--the average number of vertices per unit boundary length, where the boundary length includes both the outer ring and all interior holes. 
Higher density corresponds to more jagged or fragmented boundaries, whereas lower density indicates smoother shapes. 
This measure clearly separates smooth water bodies from highly irregular unvegetated areas and road networks.

\begin{table}[ht]
\centering
\caption{Per-class polygon complexity statistics on the training set. 
``\#Inst.'': number of polygon instances; 
``Med.~Vert.''/``$P_{90}$ Vert.''/``Max Vert.'': vertex count statistics; 
``Vert.~Density'': mean vertex count per unit perimeter.}
\label{tab:stats_poly_complexity}
\resizebox{\columnwidth}{!}{
\begin{tabular}{l|cccccc}
\toprule
Class & \#Inst. & Med.~Vert. & $P_{90}$ Vert. & Max Vert. & \#Holes & Vert.~Density \\ 
\midrule
Buildings    & 24,329 & 5  & 17  & 176 & 133  & 0.0779 \\
Roads        & 3,557  & 14 & 222 & 708 & 8,142 & 0.0506 \\
Unvegetated  & 13,254 & 8  & 54  & 454 & 8,690 & 0.0828 \\
Vegetation   & 19,324 & 7  & 25  & 314 & 1,100 & 0.0676 \\
Water        & 2,880  & 9  & 33  & 318 & 18    & 0.0461 \\
\bottomrule
\end{tabular}}
\end{table}

\noindent\textbf{Geometric Diversity and Structural Complexity.}
These statistics highlight several dataset-level properties that substantially increase the difficulty of polygonal vectorization. 
First, the extreme multi-scale variation (from very small buildings to large agricultural parcels and water bodies) requires models to operate across wide spatial ranges. 
Second, thin and elongated structures such as roads are highly sensitive to small localization errors. 
Third, many polygons contain holes or nested components, introducing additional boundary loops and complex interior topology. 
Finally, intra-class shape variability is large, especially in unvegetated and vegetation regions. 
Together, these characteristics reflect the heterogeneous nature of real-world cadastral geometry and make Deventer-512 a challenging and diverse benchmark for all-class polygonal vectorization. 

Representative visual examples are provided in Sec.~\ref{sec:sup_more_figs}.


\subsection{Challenging Visual Conditions and Subsets}
\label{sec:sup_challengingvisual}
Aerial imagery often contains regions where polygon boundaries cannot be reliably inferred from visual evidence.  
We identify two major types of challenging cases and annotate corresponding subsets on the \emph{test set}:  
(i) \emph{shadow/occlusion}, and  
(ii) \emph{semantically ambiguous boundaries}. 
Table~\ref{tab:challenging_subsets} summarizes the number of test patches belonging to each subset; 
a patch may contain multiple tags.

\vspace{4pt}

\noindent\textbf{Shadow / Occlusion.}
Targets are partially covered by trees or built structures, obscuring true boundaries and creating visually missing segments.

\vspace{4pt}

\noindent\textbf{Semantically Ambiguous Boundaries.}
A distinctive challenge of Deventer-512 is that polygons follow \emph{cadastral or administrative units} rather than purely visual contours.  
This produces three typical forms of semantic--visual discrepancy:
(1) parcel- or property-based divisions whose boundaries are not clear (\eg, private courtyards with invisible boundaries), 
(2) functional zoning rules that assign different semantic classes to visually identical surfaces (\eg, internal access roads labeled as unvegetated area), and 
(3) seasonal hydrological and vegetation changes that alter the visible extent of water bodies (\eg vegetation overgrowth or water-level fluctuations exposing sandbars). 
These cases lack reliable visual cues and make boundary localization inherently ambiguous. 
In contrast, existing land-cover datasets such as LoveDA~\cite{wang2021loveda} or ISPRS Vaihingen/Potsdam~\cite{ISPRS_Vaihingen_2D, ISPRS_Potsdam_2D} primarily follow visual boundaries.

\begin{figure}[ht]
\centering
\begin{subfigure}[t]{\linewidth}
    \centering
    \includegraphics[width=\linewidth, trim={107mm, 130mm, 107mm, 0mm}, clip]{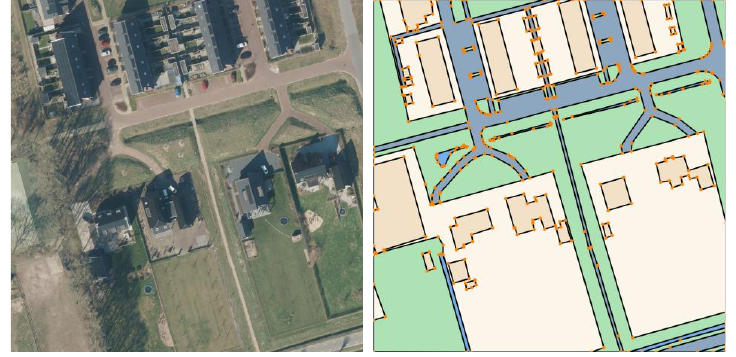}
    \caption{Semantically defined but visually invisible boundary}
\end{subfigure}
\vspace{2pt}
\begin{subfigure}[t]{\linewidth}
    \centering
    \includegraphics[width=\linewidth, trim={107mm, 69mm, 107mm, 60mm}, clip]{sec/figs/cadastral_examples.pdf}
    \caption{Semantic zoning contradicting visually identical surfaces}
\end{subfigure}
\vspace{2pt}
\begin{subfigure}[t]{\linewidth}
    \centering
    \includegraphics[width=\linewidth, trim={107mm, 7mm, 107mm, 122mm}, clip]{sec/figs/cadastral_examples.pdf}
    \caption{Seasonal hydrological changes misaligned with semantic water labels}
\end{subfigure}
\caption{
Representative examples of semantically ambiguous boundaries caused by cadastral or administrative labeling rules.  
Each row corresponds to one discrepancy type described in Sec.~\ref{sec:sup_challengingvisual}.}
\label{fig:cadastral_examples}
\end{figure}

Representative examples are shown in Fig.~\ref{fig:cadastral_examples}. 

\vspace{4pt}

\begin{table}[h]
\centering
\caption{
Challenging subsets in the test set. 
Each patch may belong to multiple subsets; percentages are computed relative to the entire test set.
}
\label{tab:challenging_subsets}
\begin{tabular}{l|cc}
\toprule
Subset & \#Patches & Ratio (\%) \\
\midrule
Shadow / Occlusion     & 194 & 88.2 \\
Ambiguous Boundaries   & 94 & 42.7 \\
\bottomrule
\end{tabular}
\end{table}

\section{Evaluation Metrics}
\label{sec:eval}

We evaluate all methods using two complementary families of metrics:
(i) \emph{per-class accuracy metrics}, assessing semantic fidelity, geometric accuracy, and topological fidelity of individual polygons; and
(ii) \emph{global topology-consistency metrics}, evaluating whether the predicted polygons collectively form a valid planar partition without gaps or overlaps. 
In addition, for the ablation study (Sec.~6.3
of the main paper), we report \emph{vertex--boundary alignment and peak-shape metrics} to quantify the quality of vertex response peaks.

\subsection{Per-class Accuracy Metrics}
\label{sec:perclass_metrics}
These metrics measure how well the predicted polygons match ground-truth semantic regions, geometric boundaries, vertex usage, and connectivity structure. All metrics are computed per class and averaged over the test set.

\paragraph{Semantic fidelity.}
\emph{IoU} (Intersection over Union) measures the region-overlap between predicted and ground-truth semantic masks (range $[0,1]$, higher is better). 
\emph{B-IoU} (Boundary IoU)~\cite{cheng2021boundary} computes IoU within a thin band around ground-truth boundaries, emphasizing boundary-level localization accuracy (range $[0,1]$, higher is better).

\paragraph{Geometric accuracy.}
\emph{PoLiS}~\cite{avbelj2014metric} is a symmetric vertex-to-boundary distance between two polygons. 
Given matched polygons $A$ and $B$ with vertex sequences $\{a_i\}_{i=1}^{n}$ and $\{b_j\}_{j=1}^{m}$, respectively, PoLiS computes the average minimal distance from each vertex of one polygon to the boundary of the other:
\begin{equation}
\begin{aligned}
d_{\mathrm{PoLiS}}(A,B)
= \tfrac{1}{2}\biggl(
    \tfrac{1}{n}\sum_{i=1}^{n} \mathrm{dist}(a_i,\partial B)
    \\ +\tfrac{1}{m}\sum_{j=1}^{m} \mathrm{dist}(b_j,\partial A)
\biggr),
\end{aligned}
\end{equation}
where $\partial A$ and $\partial B$ denote the polygon boundaries,
and $\mathrm{dist}(a_i,\partial B)$ (resp.\ $\mathrm{dist}(b_j,\partial A)$)
is the minimal Euclidean distance from $a_i$ (resp.\ $b_j$) to the edges of $B$ (resp.\ $A$). 
Lower values indicate smaller boundary displacement between the two polygons. 

\emph{MTA} (Max Tangent Angle Error)~\cite{girard2021polygonal} measures tangent-direction deviation between predicted and ground-truth contours. 
For each predicted contour $C_p$, tangent angles $\theta_p(x_k)$ are estimated at uniformly sampled boundary points $x_k$ and compared to the tangent $\theta_g(y_k)$ at the closest ground-truth point $y_k$.  
The contour-level error is the maximum angular deviation:
\begin{equation}
\mathrm{MTA}(C_p, C_g)
= \max_{k}\,
\bigl|
\mathrm{wrap}\!\left(
\theta_p(x_k) - \theta_g(y_k)
\right)
\bigr|,
\end{equation}
where $\mathrm{wrap}(\cdot)$ maps angle differences to $[-\pi,\pi]$. 
Lower values indicate better alignment of boundary tangent directions. 

Both metrics are computed on matched polygon pairs, where a prediction is matched to the ground-truth polygon with IoU exceeding 0.5 (following HiSup~\cite{xu2023hisup}). 
For each metric, we report the mean value over all matched pairs in the test set.

\paragraph{Vertex efficiency.}
\emph{$N$-ratio} measures the relative vertex usage between prediction and ground truth.  
For a matched pair $(A,B)$ with $N_A$ and $N_B$ vertices,
\begin{equation}
    \mathrm{N\text{-}ratio}(A,B)=\frac{N_A}{N_B},
\end{equation}
with values $\to 1$ being optimal: values $>1$ indicate redundant vertices, while values $<1$ indicate under-sampling. 

\emph{C-IoU} (Efficiency-aware IoU)~\cite{zorzi2022polyworld} jointly evaluates region overlap and vertex-count efficiency.  
Let $\mathrm{IoU}(A,B)$ denote the region IoU.  
The relative vertex-count deviation is defined as
\begin{equation}
    \mathrm{RD}(A,B)
    = \frac{|N_A - N_B|}{N_A + N_B}.
\end{equation}
The efficiency-aware IoU is then
\begin{equation}
    \mathrm{C\text{-}IoU}(A,B)
    = \mathrm{IoU}(A,B)\,\bigl(1 - \mathrm{RD}(A,B)\bigr).
\end{equation}
Higher values indicate both accurate region overlap and efficient vertex usage, penalizing polygons that are oversimplified or contain redundant vertices.

\paragraph{Topological fidelity.}
\emph{APLS} (Average Path Length Similarity)~\cite{van2018spacenet} is used to assess graph-level connectivity for classes
with elongated polygonal structures, namely roads and water. Since our outputs
are polygon masks rather than explicit graphs, we first convert each class mask
into a pseudo centerline graph following the skeletonization strategy of \cite{jiao2025ldpoly}. 
Concretely, we compute a 1-pixel-wide medial-axis
skeleton from the binary mask; each skeleton pixel is treated as a graph node,
and edges are inserted between 8-neighboring pixels, yielding a densely
connected pseudo graph. To reduce redundancy and focus on semantically
meaningful paths, we select all nodes with degree $\neq 2$ (endpoints and
junctions) as control nodes and decompose the graph into simple polylines by
tracing shortest paths between each pair of control nodes.

Given the ground-truth and predicted graphs $G$ and $G'$ constructed in this
way, let $\mathcal{P}$ be the set of control-node pairs $(a,b)$ in $G$ with
finite shortest-path length $L_G(a,b)$. For each $(a,b)\in\mathcal{P}$, we
locate the nearest nodes $a',b'$ in $G'$ and define the normalized path-length
discrepancy~\cite{van2018spacenet}
\begin{equation}
    \delta(a,b) = \min\!\left(1,\,
\frac{\bigl|L_G(a,b) - L_{G'}(a',b')\bigr|}{L_G(a,b)}
\right).
\end{equation}
The APLS score is then
\begin{equation}
    \mathrm{APLS}(G,G')
    = 1 - \frac{1}{|\mathcal{P}|}\sum_{(a,b)\in\mathcal{P}} \delta(a,b),
\end{equation}
ranging from $0$ (poor) to $1$ (perfect connectivity and path lengths) for the
road and water graphs.

\emph{$\chi$-err.}
Betti numbers are fundamental topological invariants that characterize
$k$-dimensional “holes’’ in a topological space~\cite{wasserman2018topological}. 
For a binary mask $M$, the $0$-th and $1$-st Betti numbers, $\beta_0(M)$ and
$\beta_1(M)$, correspond to the number of connected components and interior holes. 
The Euler characteristic is a classical topological invariant that summarizes
the global structure of a space by combining its Betti numbers through an
alternating sum~\cite{richardson2014efficient}. 
Following ~\cite{li2025topology}, we define the 2D specialization of the Euler characteristic as
\begin{equation}
\chi(M)=\beta_0(M)-\beta_1(M),
\end{equation}
and compute the Euler-characteristic error
\begin{equation}
    \chi\text{-err}(A,B)=\bigl|\chi(A)-\chi(B)\bigr|.
\end{equation}
Lower values indicate that the predicted mask preserves the correct overall topological structure in terms of the aggregate balance between connected components and interior holes.

\emph{$\beta$-err.}
Betti-number discrepancies are a standard way to quantify topological errors in image segmentation~\cite{clough2020topological}. 
Using the Betti numbers defined above, the Betti-number error between prediction $A$ and
ground truth $B$ is
\begin{equation}
    \beta\text{-err}(A,B)
    = \bigl|\beta_0(A)-\beta_0(B)\bigr|
    + \bigl|\beta_1(A)-\beta_1(B)\bigr|.
\end{equation}
Lower values indicate more faithful preservation of both connectivity
($\beta_0$) and interior-hole structure ($\beta_1$).

\subsection{Global Topology-Consistency Metrics}
\label{sec:global_topology}
These metrics assess whether the predicted polygons collectively form a topologically valid planar partition of the image domain. 

\paragraph{Gap rate.}
The fraction of the image domain not covered by any predicted polygon. 
This captures missing regions in the partition (lower is better).

\paragraph{Inter-class overlap.}
The fraction of area simultaneously assigned to two or more different classes. 
This identifies semantic conflicts between classes (lower is better).

\paragraph{Intra-class overlap.}
The fraction of area overlapped by multiple polygons of the same class. 
This indicates boundary intersections or duplicated geometry within a single semantic region. (lower is better).

\paragraph{Shared-edge consistency.}
In a valid planar partition (a planar straight-line graph), every \emph{interior} 
edge must be incident to exactly two polygons. 
We therefore measure the fraction of predicted interior edges that satisfy this 
two-sided incidence condition:
\begin{equation}
\resizebox{0.433\textwidth}{!}{%
$\displaystyle
\mathrm{SEC}=\frac{\#\{\text{interior edges shared by exactly two polygons}\}}
     {\#\{\text{all interior edges}\}}.
     $}
\end{equation}
Edges lying on the patch boundary are excluded. 
Higher values indicate stronger adherence to planar-partition topology.

\subsection{Vertex--Boundary Alignment and Peak-Shape Metrics}
\label{sec:sup_v2b}
In the ablation study (Sec.~6.3 of the main paper)
we further analyze the quality of vertex response peaks. 
Let $\mathcal{V}$ be the set of predicted vertices, obtained by non-maximum suppression (NMS) on the vertex heatmap, and let $\mathcal{B}_c$ denote the predicted semantic boundaries of all classes (extracted as class-transition boundaries from the multi-class segmentation mask). 

\paragraph{Vertex-to-boundary alignment (V2B@$ \delta $).}
We measure the fraction of predicted vertices that lie within a distance
$\delta$ of their nearest predicted semantic boundaries. 
For each vertex $v \in \mathcal{V}$, we compute its minimal Euclidean distance to the predicted boundary, 
\begin{equation}
    d(v)=\min_{x\in\hat{\mathcal{B}}_{c(v)}} \mathrm{dist}(v,x),
\end{equation}
and define
\begin{equation}
\mathrm{V2B}_\delta
=
\frac{
\#\{\,v \in \mathcal{V} \mid d(v) \le \delta\,\}
}{
|\mathcal{V}|
}.
\end{equation}
$\mathrm{V2B}_\delta \in [0,1]$ (higher is better) quantifies how well the
predicted vertex peaks align with the predicted class boundaries, a property
crucial for reliable topological reconstruction. 

\paragraph{Peak-shape metrics.}
For each detected peak location $x_0$, we extract a local $K \times K$ patch $H$
from the corresponding vertex heatmap and normalize it so that $H(x_0)=1$. 
We then characterize the local peak shape using three metrics commonly used in imaging and spectroscopy to quantify peak width and sharpness. 

\emph{Area@0.5.}
We first threshold the patch at half maximum and consider the 8-connected
component that contains the peak center $x_0$.
Let $\mathcal{C}_{0.5} = \{x \in H \mid H(x) \ge 0.5\}$ be this component.
We define
\begin{equation}
\mathrm{Area@0.5}
=
\#\,\mathcal{C}_{0.5},
\end{equation}
i.e., the number of pixels in the central half-maximum support
(smaller values indicate more spatially concentrated peaks).
Since the patch size $K\times K$ is fixed, areas are directly comparable across methods.

\emph{FWHM.}
To capture peak width and anisotropy, we measure axis-aligned full widths at half maximum along the horizontal and vertical profiles through $x_0$. 
Let $p_x$ and $p_y$ be the 1D profiles of $H$ along the central row and column,
respectively.
We define $\mathrm{FWHM}_x$ (resp. $\mathrm{FWHM}_y$) as the distance between the two points where $p_x$ (resp. $p_y$) crosses $0.5$, using linear interpolation between neighboring pixels.
Lower values correspond to narrower peaks.

\emph{Sharpness.}
To quantify local curvature at the peak, we apply Gaussian smoothing with
standard deviation $\sigma$ to $H$ and compute the Hessian at $x_0$.
Let $\Delta H(x_0)$ denote the Laplacian of the smoothed patch at the center.
We define a scale-normalized sharpness measure
\begin{equation}
\mathrm{Sharpness}
=
\max\bigl(0,\,-\sigma^2 \Delta H(x_0)\bigr),
\end{equation}
so that sharper, more peaked responses yield higher values.

All peak-shape metrics are averaged over all detected peaks and are used only in
our ablation analysis.

\section{Method: Additional Details}
\label{sec:sup_method}

\subsection{SSC: Architecture Details}
\label{sec:sup_ssc_arch}

\paragraph{Semantic encoder.}
We instantiate $S_\psi$ using a VMamba-based state-space backbone \cite{liu2024vmamba} operating at four spatial scales. 
Its multi-level features are unified by a UPerNet-style pyramid aggregation module~\cite{xiao2018unified}, producing
a single semantic feature map $S_\psi(I)\in\mathbb{R}^{C_s\times \tfrac{H}{4}\times \tfrac{W}{4}}$ aligned to the quarter-resolution grid of the image. 

\paragraph{Segmentation head for explicit supervision.}
To provide direct supervision for $S_\psi$, we attach a lightweight segmentation head composed of two submodules. 
The first is a stack of three $3{\times}3$ Conv–BN–ReLU blocks that refine the semantic features. 
The second is a predictor consisting of a $3{\times}3$ convolution followed by a $1{\times}1$ projection. 
The output is then bilinearly upsampled to full resolution to produce per-pixel logits. 

\paragraph{Conditioning fusion.}
The semantic feature map $S_\psi(I)$ is injected into the denoising network through channel-wise concatenation at the quarter-resolution latent scale. 
Empirically, cross-attention offered negligible improvement while incurring substantial computational overhead, so concatenation is adopted for all experiments.

\paragraph{Denoising UNet.}
Our denoiser follows the widely used UNet architecture employed in latent diffusion frameworks~\cite{rombach2022high}, but is substantially lightened to suit the ACPV setting. 
We use $128$ base channels with channel multipliers $[1,2,2]$, a single residual block per resolution stage, and self-attention layers at resolutions $32{\times}32$ and $16{\times}16$ (attention resolutions $4$ and $8$ in the latent grid), with $32$ head channels. 
The network receives a $640$-channel input formed by concatenating the noisy latent vector with $S_\psi(I)$ and predicts the denoising target in the latent diffusion formulation. 
Despite being substantially reduced in depth and width, this lightweight denoiser provides sufficient capacity for the ACPV task, and deeper variants did not yield meaningful improvements in our experiments.

\paragraph{Latent Distributional Vertex Modeling}
\label{sec:sup_latent}
We use the publicly released, pretrained KL-regularized variational autoencoder (VAE) employed in latent diffusion frameworks
\cite{rombach2022high}. 
It maps an $H{\times}W$ heatmap to a $\tfrac{H}{4}\times \tfrac{W}{4}\times 4$ latent vector. 
Both the encoder and decoder remain frozen and simply provide a fixed mapping between pixel space and the latent space used for diffusion. 
All results reported in the paper use this default KL-VAE without modification.

\subsection{PSLG Construction and Vertex-Guided Simplification}
To further clarify the PSLG-based polygonization pipeline described in Sec.~4.2 of the main paper, we provide a more detailed explanation together with an illustrative example.

The planar straight-line graph (PSLG) used in ACPV is constructed from semantic label transitions in the segmentation mask. 
Pixels where semantic labels change are treated as vertices, and adjacent transitions on the image grid are connected to form edges, forming a planar graph aligned with semantic boundaries.

Redundant vertices are then removed through PSLG simplification guided by the predicted vertex heatmaps. 
The heatmaps provide sparse cues for selecting representative vertices along the PSLG using distance-based criteria, reducing vertex redundancy while preserving geometrically meaningful boundary points.

Polygon connectivity is subsequently obtained by graph traversal on the simplified PSLG rather than by directly linking predicted vertices, producing consistent polygon boundaries derived from the planar graph.

This PSLG-based polygonization is also more robust to imperfect segmentation masks. 
Segmentation predictions may contain artifacts such as blobby regions, fragmented areas, or spurious holes. 
Conventional polygonization methods (e.g., raster contour tracing followed by Douglas--Peucker simplification) tend to propagate such raster noise into polygon boundaries. 
In contrast, vertex-guided PSLG simplification suppresses these artifacts by selecting geometrically meaningful vertices on the planar graph.

\begin{figure}[t]
\centering
\includegraphics[width=\linewidth, trim=44pt 5pt 140pt 142pt, clip]{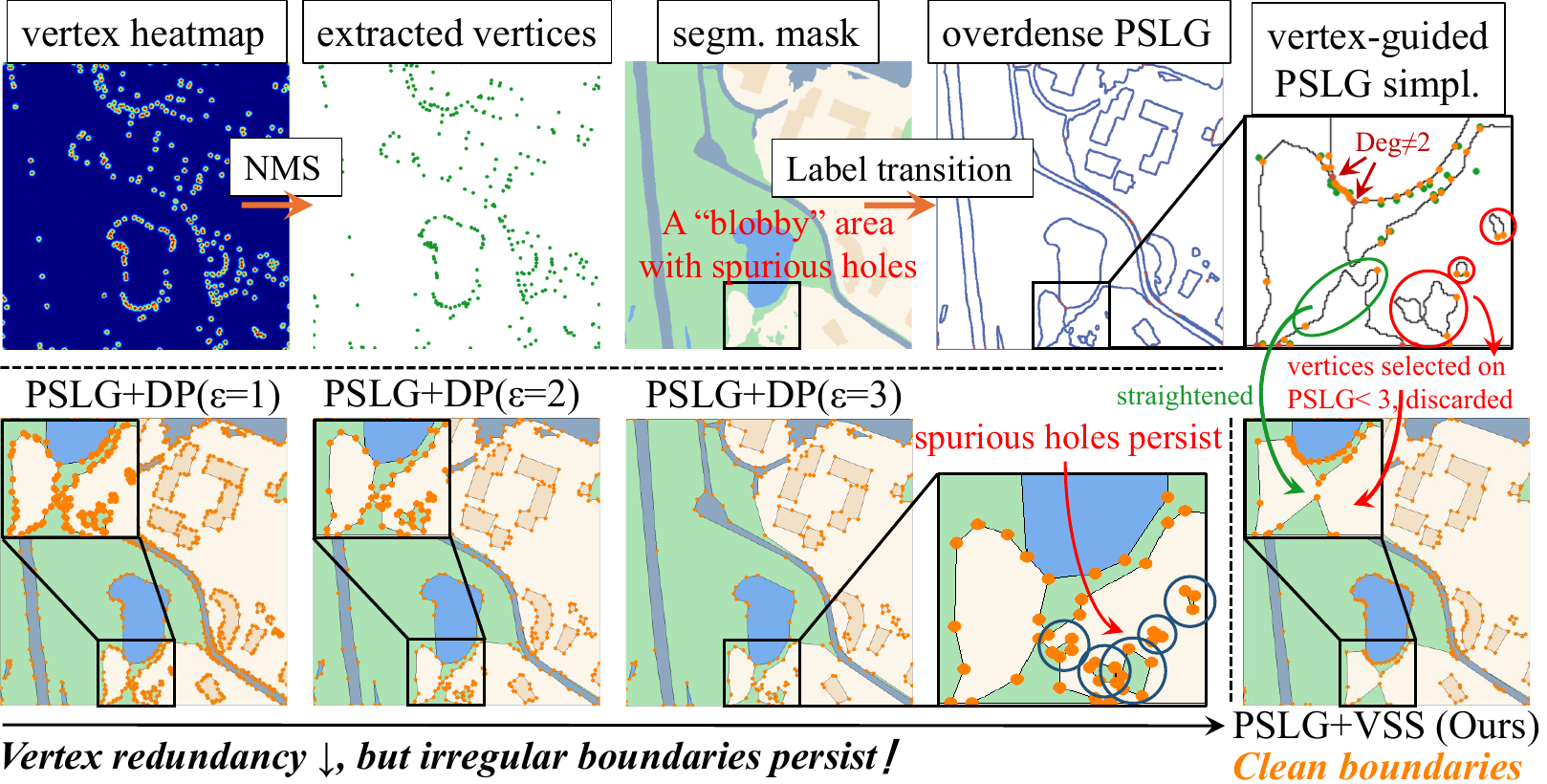}
\caption{Illustration of the PSLG-based polygonization pipeline under segmentation noise.}
\label{fig:R1_pipeline}
\end{figure}

Figure~\ref{fig:R1_pipeline} illustrates the complete pipeline under segmentation noise and compares the proposed approach with DP-based contour tracing. 
The PSLG-based polygonization yields cleaner polygon boundaries and more stable vertex selection under noisy segmentation conditions.

\subsection{Topological Reconstruction: Sufficient Condition Proof}
\label{sec:sup_proof}
For convenience, we restate Proposition~1
from the main paper and provide a more detailed justification. 
\setcounter{proposition}{0}  
\begin{proposition}[Sufficient condition for ACPV compliance]
Let $G=(V,E)$ be a planar straight-line graph embedded in $\mathbb{R}^2$.
Suppose that:
\begin{propenum}
    \item Each edge $e \in E$ lies along a label-transition boundary in the
    multi-class mask $\hat{M}$, and the two faces adjacent to $e$ have distinct
    semantic labels; and
    \label{con1app}
    \item The vertex set $V$ consists exclusively of \emph{geometric
    keypoints}, i.e., vertices whose removal would alter the graph’s topology
    or geometric structure (junctions, salient corners, and endpoints of
    semantic transitions).
    \label{con2app}
\end{propenum}
Then the polygonal partition obtained by tracing all face boundaries on $G$ and
assigning face labels according to $\hat{M}$ satisfies all ACPV constraints
(a)–(f).
\end{proposition}

\noindent\textbf{Proof sketch.}
Because $G$ is a planar straight-line graph whose embedding includes the image boundary, it induces a planar subdivision $\mathcal{F}$ of the domain into finitely many bounded faces with pairwise–disjoint interiors.  
Tracing the boundary cycle of each face in $\mathcal{F}$ yields a simple polygon.

Condition~\ref{con1app} guarantees that every edge of $G$ coincides with a semantic transition in $\hat{M}$ and is incident to two faces with distinct labels. 
Hence each face in $\mathcal{F}$ admits a unique semantic label compatible with all its boundary edges, producing a well-defined face labeling. 
Because the interiors of the faces are disjoint and their union equals the whole domain, the induced face labeling defines a gap-free and overlap-free partition.

Condition~\ref{con2app} requires that the vertex set $V$ contains all geometric keypoints, thereby preventing spurious degree-two vertices or self-intersecting boundary cycles. 
Consequently, each edge is shared by at most two faces with opposite orientations, implying consistent shared boundaries across classes.

Together, these properties imply that the labeled face decomposition of $G$ forms a valid all-class planar partition satisfying all ACPV constraints (a)–(f).

\section{Implementation and Training Details}
\label{sec:sup_impl}
This section summarizes all implementation details for ACPV-Net and the baseline methods, including dataset preprocessing, training schedules, hyperparameters, and fairness considerations. 
Our main experiments are conducted on two benchmarks, Deventer-512 and WHU-Building. 
All ablation studies are run on Deventer-512. 

\subsection{Datasets and Preprocessing}
\label{sec:sup_data}
All models operate on fixed-size $512{\times}512$ patches. 
For Deventer-512, we directly use the official $512{\times}512$ patches and training/validation/test splits (Sec.~5
of the main paper), which contain 1716/219/220 patches with 5 semantic classes. 
For WHU-Building, we adopt the official split on the original images and then extract non-overlapping $512{\times}512$ patches, discarding patches without buildings, resulting in 9344/4015/5388 patches for training/validation/test. 

For ACPV-Net, we use low-intensity geometric augmentation only: on Deventer-512 we apply random horizontal/vertical flips, and on WHU-Building we apply random horizontal/vertical flips plus random $90^\circ$ rotations. 
For all baselines, we keep the authors’ official implementations and their default augmentation pipelines, which typically include random flips and rotations and, for some methods, additional photometric jitter or stronger geometric transforms. 

\subsection{ACPV-Net Training Details}
\label{sec:sup_acpv_details}
Training configurations differ slightly between Deventer-512 and WHU-Building due to dataset size. 
Unless otherwise stated, DDIM with $T{=}200$ steps is used for inference, and an exponential moving average (EMA) with decay $0.9999$ is applied to all model weights.

\paragraph{Deventer-512.}
We train ACPV-Net for $160{,}000$ iterations with batch size $8$. 
Optimization uses AdamW with learning rate $6{\times}10^{-5}$, $(\beta_1,\beta_2){=}(0.9,0.999)$, and weight decay $0.01$. 
A linear warmup of $1{,}500$ iterations is followed by a PolyLR schedule with power $1.0$ and minimum learning rate $0$. 

\paragraph{WHU-Building.}
Due to the larger training set, ACPV-Net is trained for $869{,}000$ iterations with batch size $8$. 
We use the same AdamW configuration as above. 
Learning rate warmup lasts $8{,}150$ iterations, followed by the same PolyLR decay. 

\subsection{Baselines: Training Details}
\label{sec:sup_baselines}
We evaluate five state-of-the-art polygonization methods: DeepSnake~\cite{peng2020deep}, FFL~\cite{girard2021polygonal},
TopDiG~\cite{yang2023topdig}, HiSup~\cite{xu2023hisup}, and GCP~\cite{zhang2025global}. 
All models are trained from scratch on our datasets using the authors' official public implementations. 
We strictly follow their recommended hyperparameters, optimization settings, and training schedules. 
No architectural modifications are made.

\paragraph{DeepSnake.}
We use the publicly released implementation \cite{liu2021dance} and train the model for 100 epochs on both Deventer-512 and WHU-Building. 
Unlike the other baselines, DeepSnake supports multi-class training and is therefore trained once on the full Deventer-512 dataset rather than per semantic class.

\paragraph{FFL.}
Following the official implementation, we train each model for 1,000 epochs and select the checkpoint with the best validation performance. 
We apply this setting separately to every semantic class in Deventer-512 and to the WHU-Building dataset. 
For polygonization, we use the Active Skeleton Model (ASM) as recommended in the original paper.

\paragraph{TopDiG.}
We follow the official two-stage training protocol: 
(1) training the node-detection network, followed by 
(2) training the full TopDiG model, including the adjacency-prediction modules. 
Both stages use the authors' early-stopping strategy exactly as implemented in the public code.

A key task-specific hyperparameter is the maximum number of nodes allowed per image. 
For Deventer-512, we set this value per semantic class to 600 (buildings), 500 (roads), 700 (unvegetated), 400 (vegetation), and  
200 (water), each chosen to cover the vast majority of image patches within its semantic class. 
For water bodies, a limit of 100 covers most patches but yields poor performance; we therefore use 200, which results in normal behavior for this class. 
For WHU-Building, we use a node limit of 300.

\paragraph{HiSup.}
Following the official implementation, HiSup is trained for 30 epochs without
early stopping on every semantic class of Deventer-512 and WHU-Building.

\paragraph{GCP.}
Following the official implementation, GCP is trained in two stages.  
(1) The Mask2Former backbone is trained for 50 epochs as the base instance segmentation model.  
(2) The polygonization modules are then trained for 12 epochs using the recommended number of queries ($Q{=}300$).  
The same two-stage schedule is applied to every semantic class of Deventer-512 and WHU-Building. 

\paragraph{Fairness.}
For every baseline, we keep the original optimization settings (learning rate, scheduler, batch size, optimizer, augmentation) as released by the authors. 
This ensures that each method is evaluated under its intended operating regime without re-tuning or weakening.


\begin{figure*}[t!]
\centering
\begin{subfigure}[t]{\linewidth}
    \centering
    \includegraphics[width=0.93\linewidth, trim={0mm, 60mm, 0mm, 60mm}, clip]{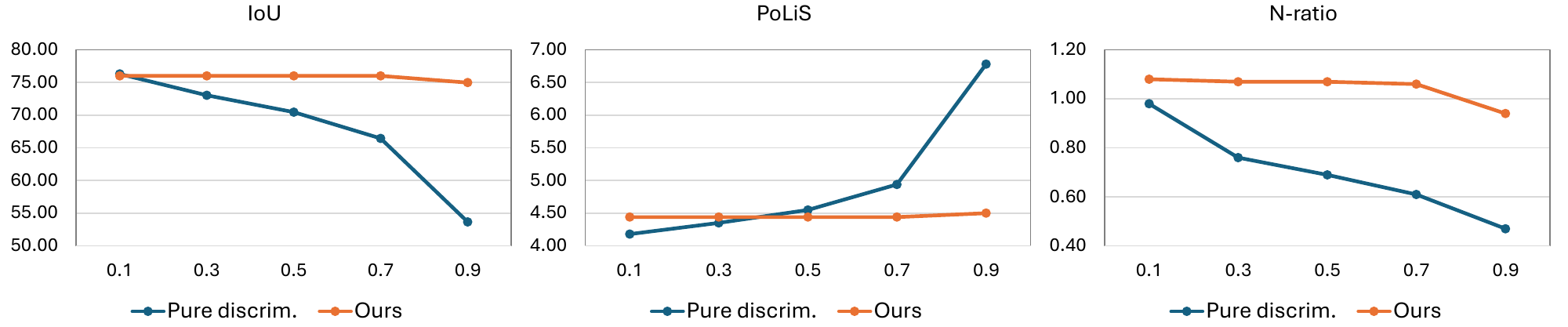}
    \caption{Road}
\end{subfigure}
\vspace{2pt}
\begin{subfigure}[t]{\linewidth}
    \centering
    \includegraphics[width=0.93\linewidth, trim={0mm, 60mm, 0mm, 55mm}, clip]{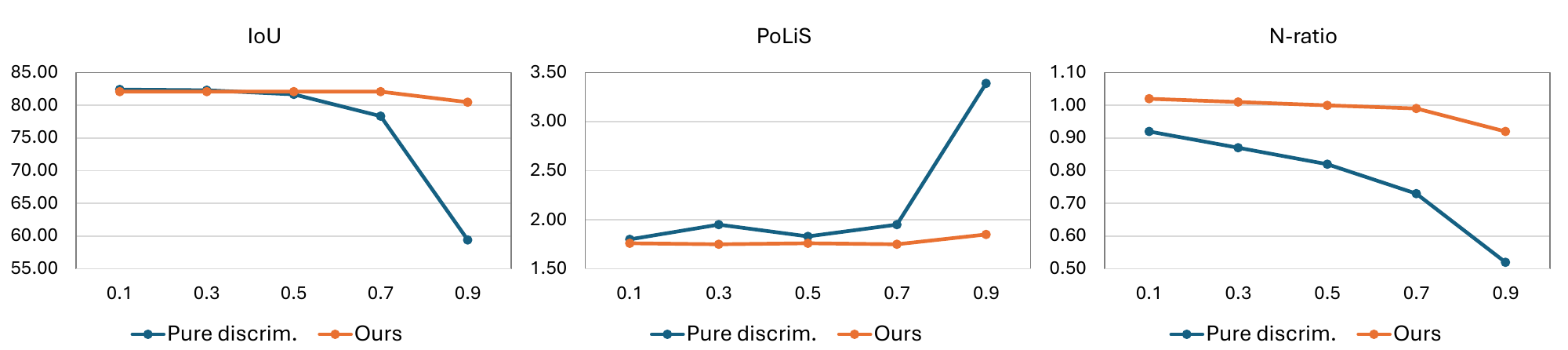}
    \caption{Building}
\end{subfigure}
\vspace{2pt}
\begin{subfigure}[t]{\linewidth}
    \centering
    \includegraphics[width=0.93\linewidth, trim={0mm, 60mm, 0mm, 55mm}, clip]{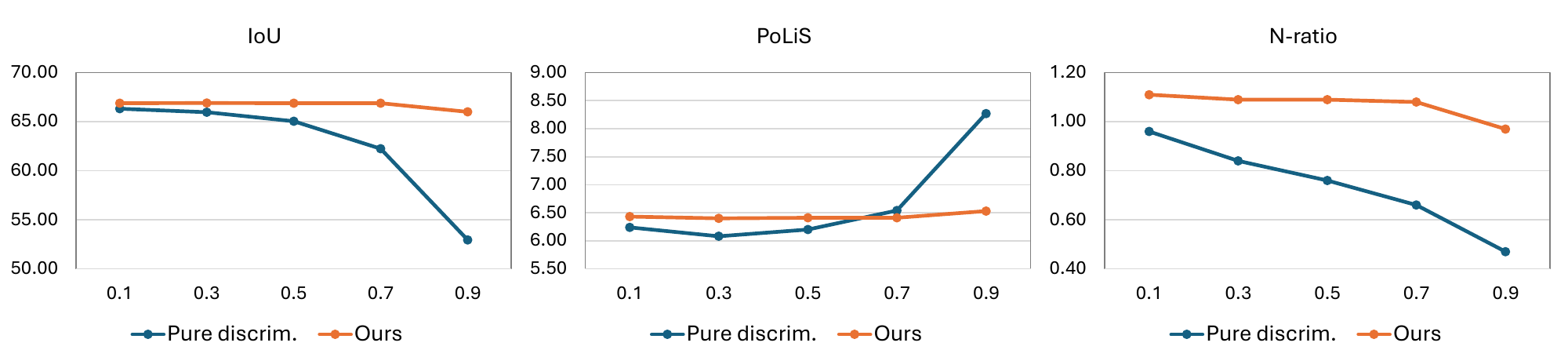}
    \caption{Unvegetated}
\end{subfigure}
\begin{subfigure}[t]{\linewidth}
    \centering
    \includegraphics[width=0.93\linewidth, trim={0mm, 60mm, 0mm, 55mm}, clip]{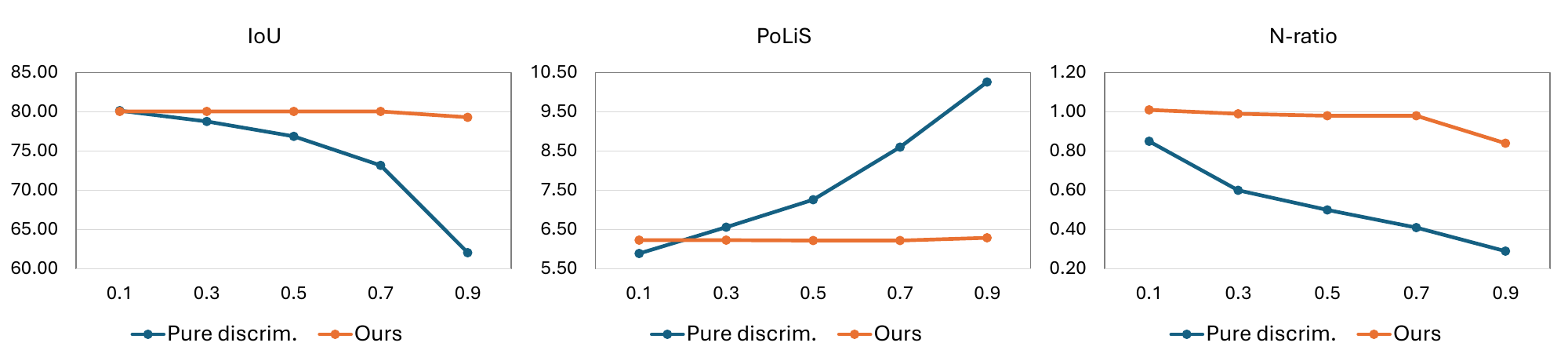}
    \caption{Vegetation}
\end{subfigure}
\begin{subfigure}[t]{\linewidth}
    \centering
    \includegraphics[width=0.93\linewidth, trim={0mm, 60mm, 0mm, 55mm}, clip]{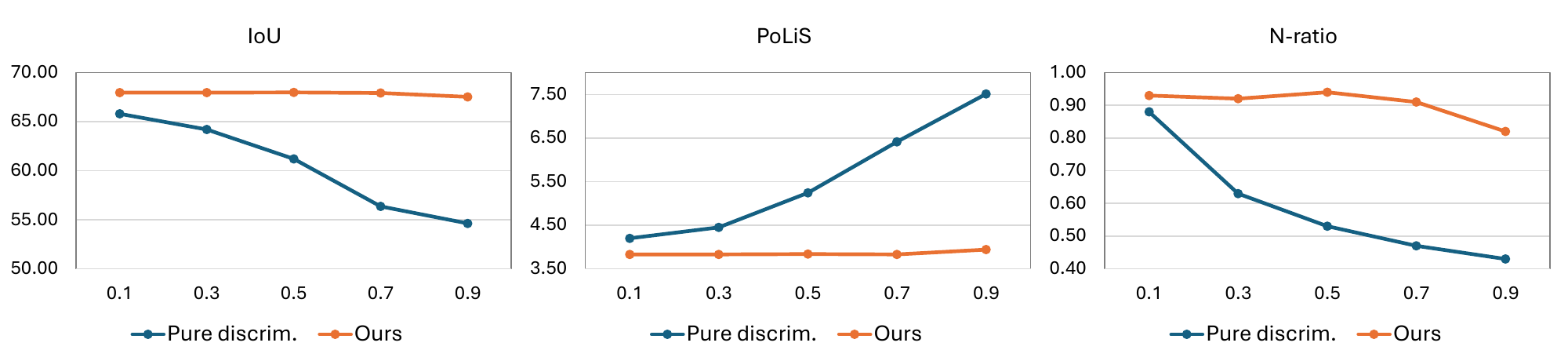}
    \caption{Water}
\end{subfigure}
\caption{Sensitivity of polygonal quality to the NMS threshold used for vertex extraction. Performance of pure discriminative decoder vs.\ our distributional vertex modeling measured in IoU (semantic fidelity, $\uparrow$ better), PoLiS (geometric accuracy, $\downarrow$ better), and N-ratio (vertex efficiency, $\rightarrow1$ better).}
\label{fig:nms_sensitivity}
\end{figure*}

\section{Extended Experiments}
\label{sec:sup_extended}

\subsection{Per-Class Topology Statistics}
\label{sec:sup_topology_per_class}
To complement the global topology evaluation in Table 7 of the main paper
we further report per-class intra-class overlap rates in Table~\ref{tab:deventer_topology_per_class}. 
While inter-class overlaps and gaps can arise when each semantic class is polygonized independently, intra-class overlaps are not expected, since these methods are specifically designed for single-class outline extraction. 
Such intra-class violations indicate topological errors within the category, such as self-intersections or other invalid polygon representations. 

ACPV-Net achieves zero measured intra-class overlap across all categories in our evaluation, indicating that its outputs are topologically consistent both within each semantic class and at the global planar-partition level.
\begin{table}[t]
  \centering
  \caption{
  \textbf{Per-class intra-class overlap rates (\%) on Deventer-512.} Best values are shown in \textbf{bold}.
  }
  \label{tab:deventer_topology_per_class}
  \renewcommand{\arraystretch}{0.9}
  \setlength{\tabcolsep}{6pt}
  \resizebox{\columnwidth}{!}{%
  \begin{tabular}{l|ccccc}
    \toprule
    Method & Building $\downarrow$ & Road $\downarrow$ & Unvegetated $\downarrow$ & Vegetation $\downarrow$ & Water $\downarrow$ \\
    \midrule
    DeepSnake~\cite{peng2020deep} & 5.92 & 12.33 & 21.14 & 28.76 & 19.55 \\
    FFL~\cite{girard2021polygonal} & \textbf{0.00} & 0.01 & 0.06 & 0.01 & \textbf{0.00} \\
    TopDiG~\cite{yang2023topdig} & \textbf{0.00} & \textbf{0.00} & \textbf{0.00} & \textbf{0.00} & \textbf{0.00} \\
    HiSup~\cite{xu2023hisup} & \textbf{0.00} & \textbf{0.00} & \textbf{0.00} & \textbf{0.00} & \textbf{0.00} \\
    GCP~\cite{zhang2025global} & 6.54 & 0.31 & 6.91 & 29.05 & 0.08 \\
    \midrule
    \cellcolor{oursbg} ACPV-Net (Ours) 
    & \cellcolor{oursbg}\textbf{0.00} 
    & \cellcolor{oursbg}\textbf{0.00} 
    & \cellcolor{oursbg}\textbf{0.00} 
    & \cellcolor{oursbg}\textbf{0.00} 
    & \cellcolor{oursbg}\textbf{0.00} \\
    \bottomrule
  \end{tabular}%
  }
\end{table}

\subsection{Robustness to NMS Thresholds}
\label{sec:sup_robustness}
To further evaluate the stability of distributional vertex modeling, we analyze the sensitivity of polygonal metrics with respect to the NMS threshold $\tau$ used for extracting vertices from the predicted heatmaps. 
The NMS kernel size is fixed to $3{\times}3$, matching the spatial scale of the Gaussian kernels used to construct ground-truth vertex heatmaps, and therefore does not constitute a tunable hyperparameter.

Figure~\ref{fig:nms_sensitivity} reports per-class sensitivity curves for the
distributional model and the pure discriminative decoder.  
The distributional model exhibits minimal variation across a wide range of NMS thresholds, whereas the discriminative baseline shows pronounced fluctuations in IoU (semantic fidelity, $\uparrow$ better), PoLiS (geometric accuracy, $\downarrow$ better), and N-ratio (vertex efficiency, $\rightarrow1$ better). 
This observation is consistent with the peak-shape analysis in 
Sec.~6.3
of the main paper: sharper and more compact peaks produced by latent diffusion lead to more stable vertex extraction and consequently more robust polygon reconstruction.

\begin{figure}[t!]
  \centering
  \includegraphics[width=\linewidth]{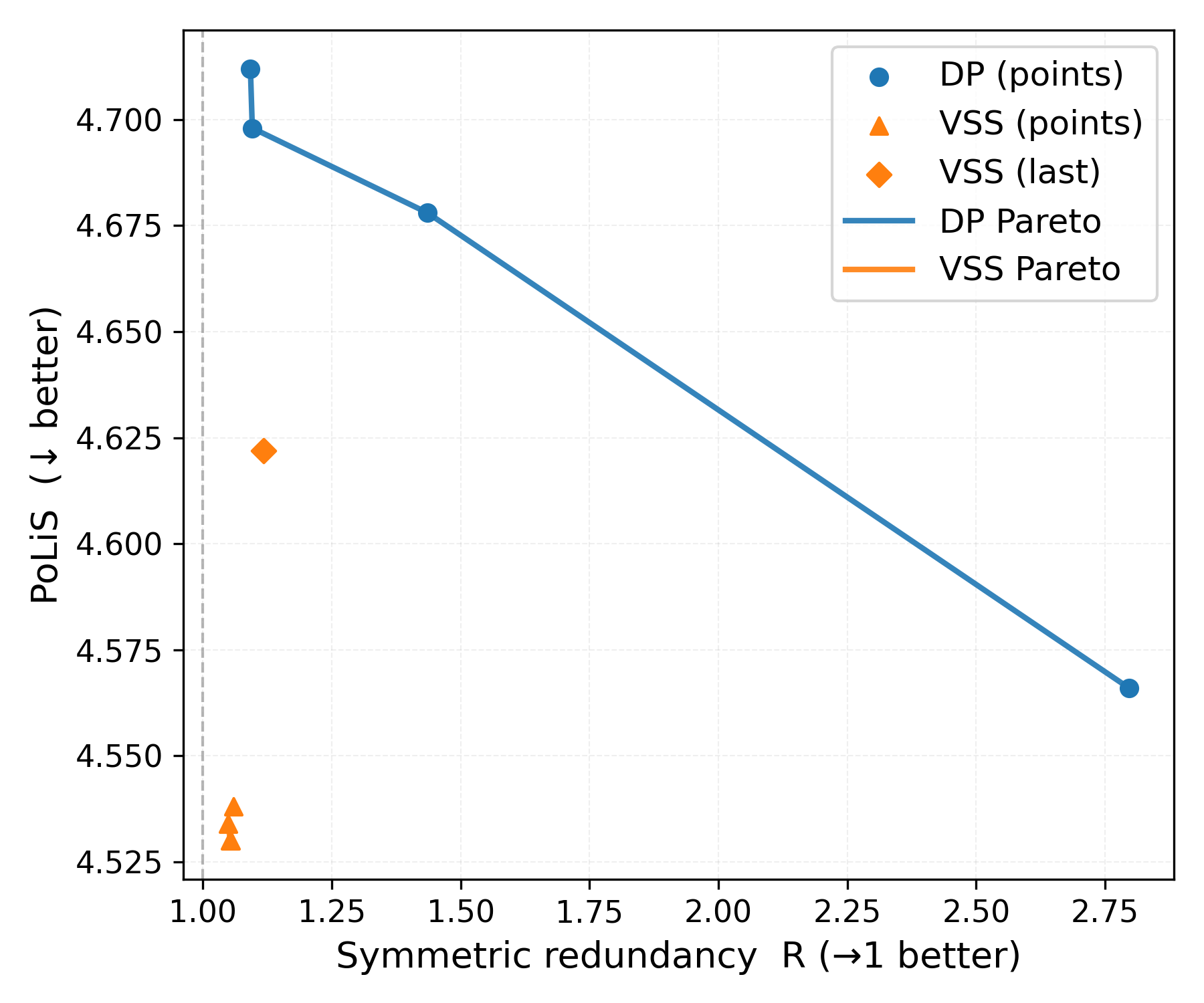}
  \caption{Class-averaged Pareto fronts between geometry error (PoLiS, $\downarrow$ better) and symmetric redundancy 
  \(R=\max(\text{N-ratio},\,1/\text{N-ratio})\) ($\rightarrow1$ better) for \emph{PSLG+DP (w/o VSS)} and \emph{PSLG+VSS (ours)}. The diamond marker denotes the extreme case ($\tau=0.9$) for VSS. }
  \label{fig:pareto_polis_nratio}
\end{figure}

\subsection{VSS vs.\ DP: Pareto Fronts}
\label{sec:sup_pareto}
To complement the discussion in
Sec. 6.3
of the main paper, we compare our vertex-guided subset selection (VSS) with the classical Douglas–Peucker (DP) simplification. 
For each method, we sweep its respective control parameter (NMS threshold for VSS; tolerance $\varepsilon$ for DP) and plot the resulting class-averaged trade-off between geometric error (PoLiS, lower is better) and symmetric redundancy $R=\max(\text{N-ratio},1/\text{N-ratio})$ (closer to 1 is better).

Figure~\ref{fig:pareto_polis_nratio} shows that VSS consistently achieves lower redundancy at comparable or lower PoLiS, indicating that learned vertex cues provide a more effective basis for keypoint preservation than purely geometric simplification. 
The diamond marker denotes the extreme case $\tau=0.9$ for VSS.

\subsection{Sanity Check for the Frozen KL-VAE Encoder--Decoder}
\begin{figure*}[t]
\centering
\begin{subfigure}[t]{\linewidth}
    \centering
    \includegraphics[width=\linewidth, trim={0mm, 60mm, 0mm, 60mm}, clip]{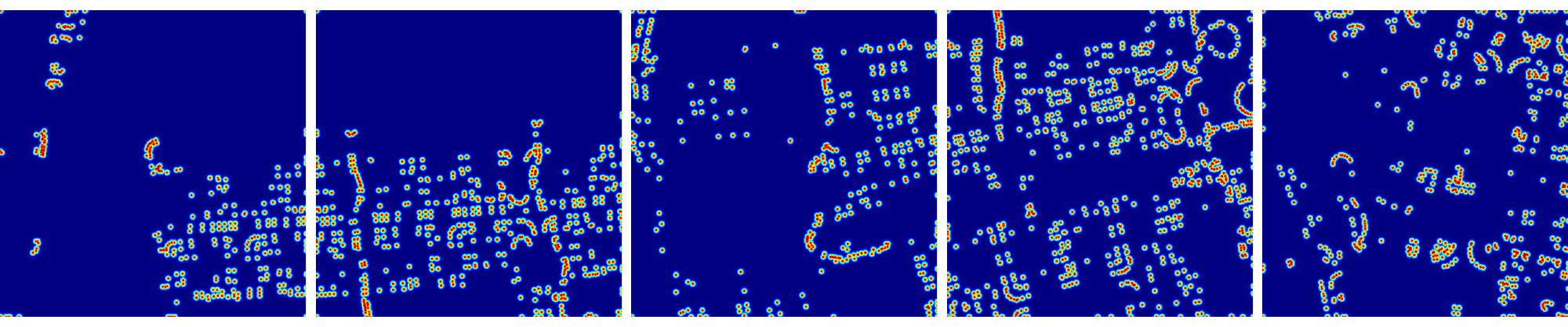}
    \caption{Reconstructed}
\end{subfigure}
\vspace{2pt}
\begin{subfigure}[t]{\linewidth}
    \centering
    \includegraphics[width=\linewidth, trim={0mm, 60mm, 0mm, 55mm}, clip]{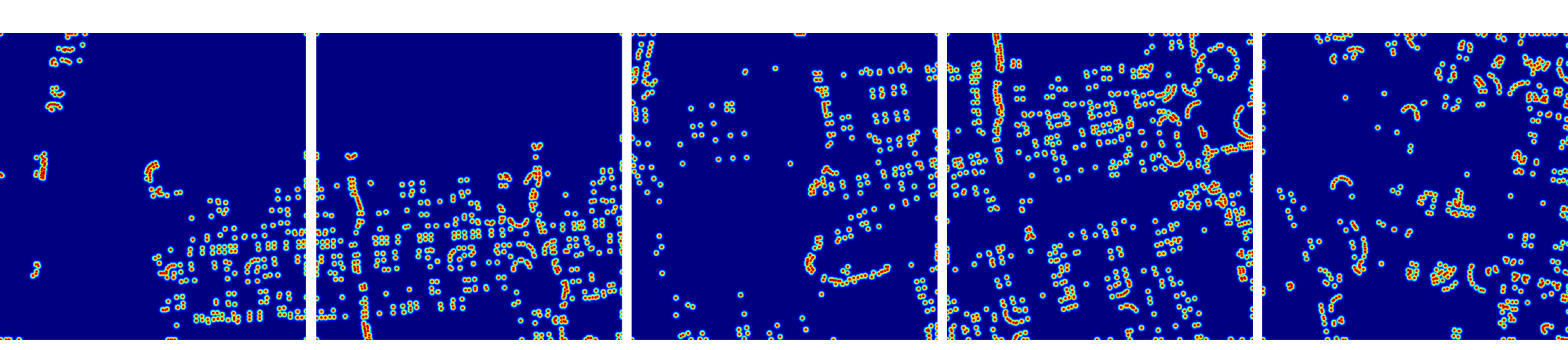}
    \caption{Ground truth}
\end{subfigure}
\caption{Sanity check for the frozen KL-VAE Encoder-Decoder. Vertex heatmaps are reconstructed with highly preserved peak location, anisotropy, and contrast.}
\label{fig:vertice_heatmap}
\end{figure*}

To ensure that using a pretrained and frozen KL-VAE does not distort the vertex heatmaps used by ACPV-Net, we perform a direct encoder–decoder reconstruction test. 
Each ground-truth vertex heatmap is passed through the frozen VAE encoder to obtain a latent representation, which is then decoded
back to the heatmap space. 
Representative examples are shown in Fig.~\ref{fig:vertice_heatmap}.

The reconstructed heatmaps faithfully preserve peak location, anisotropy, and contrast, with negligible visual degradation. 
This confirms that the frozen KL-VAE has sufficient capacity for representing vertex heatmaps and does not introduce task-relevant artifacts.

\subsection{Cross-region Generalization to External Aerial Datasets}
\label{sec:sup_crossdomain}
To assess cross-region generalization, we apply the model trained on Deventer-512 directly (no fine-tuning) to aerial imagery from multiple countries and independent national mapping agencies. Specifically, we evaluate on orthorectified RGB aerial photographs at $0.3\,\mathrm{m}$ ground resolution from (i) Eastern Tyrol and Innsbruck in western Austria, acquired through the Austrian federal aerial survey program, 
and (ii) Bellingham in Washington State, USA, obtained from local governmental aerial surveys. 
We further test on urban aerial imagery from Christchurch, New Zealand, captured by the national LINZ aerial mapping program and released at $0.3\,\mathrm{m}$ resolution. 

\begin{figure*}[t]
\centering
\begin{subfigure}[t]{\linewidth}
    \centering
    \includegraphics[width=\linewidth, trim={0mm 25mm 0mm 25mm}, clip]{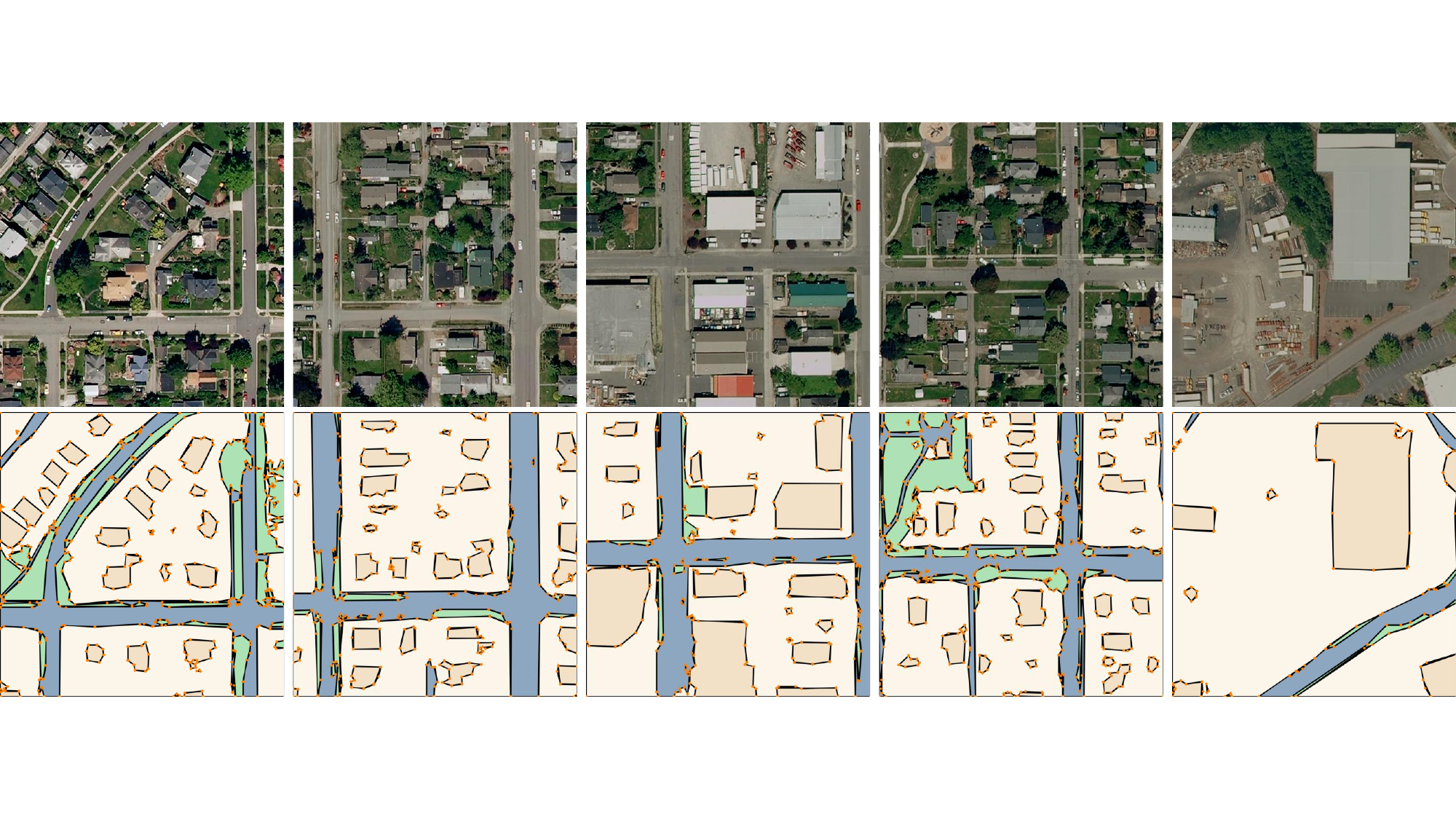}
    \caption{Bellingham (USA)}
\end{subfigure}
\vspace{4pt}
\begin{subfigure}[t]{\linewidth}
    \centering
    \includegraphics[width=\linewidth, trim={0mm 25mm 0mm 25mm}, clip]{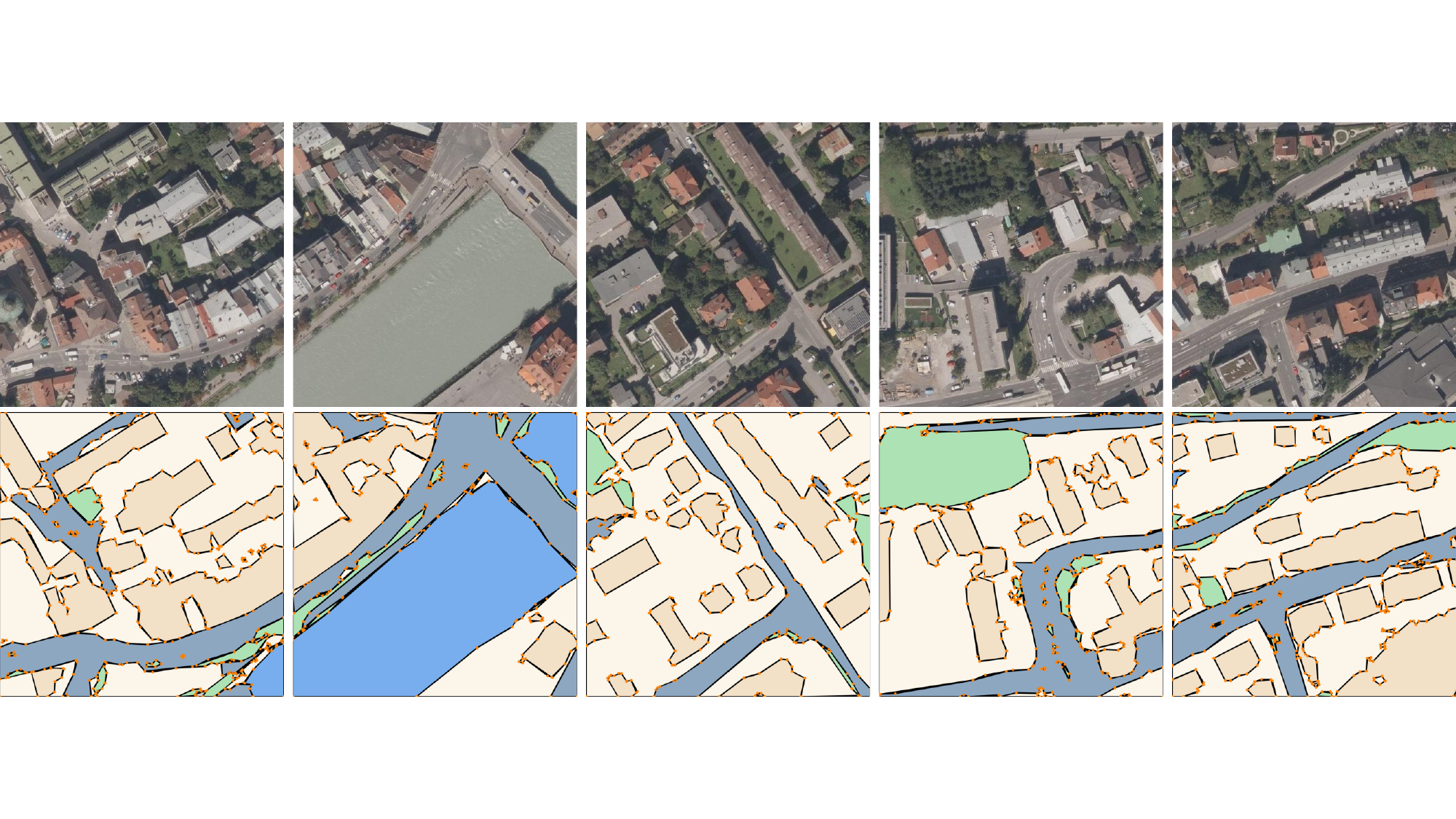}
    \caption{Innsbruck (Austria)}
\end{subfigure}

\caption{
\textbf{Cross-region qualitative results (Part~1).} 
Polygons of different semantic classes are shown in distinct colors. Predicted vertices are marked as orange points. Topological inconsistencies, if present, would appear in black (gaps) or red (overlaps). These examples illustrate ACPV-Net’s behavior on imagery from independent national mapping agencies \textbf{without fine-tuning}. 
}
\label{fig:cross_region_vis_part1}
\end{figure*}

\begin{figure*}[t]
\centering
\begin{subfigure}[t]{\linewidth}
    \centering
    \includegraphics[width=\linewidth, trim={0mm 25mm 0mm 25mm}, clip]{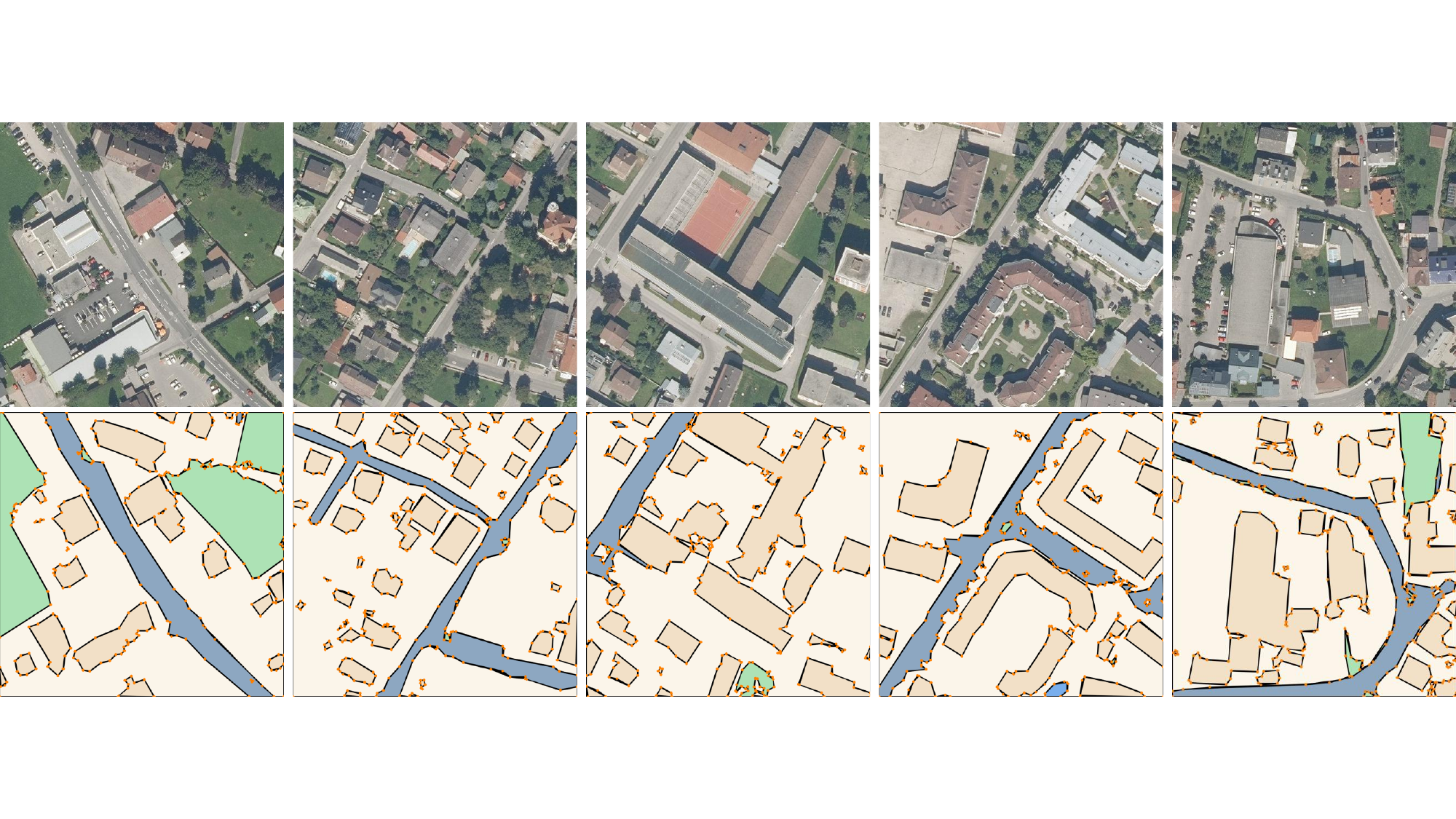}
    \caption{Eastern Tyrol (Austria)}
\end{subfigure}

\vspace{4pt}
\begin{subfigure}[t]{\linewidth}
    \centering
    \includegraphics[width=\linewidth, trim={0mm 25mm 0mm 25mm}, clip]{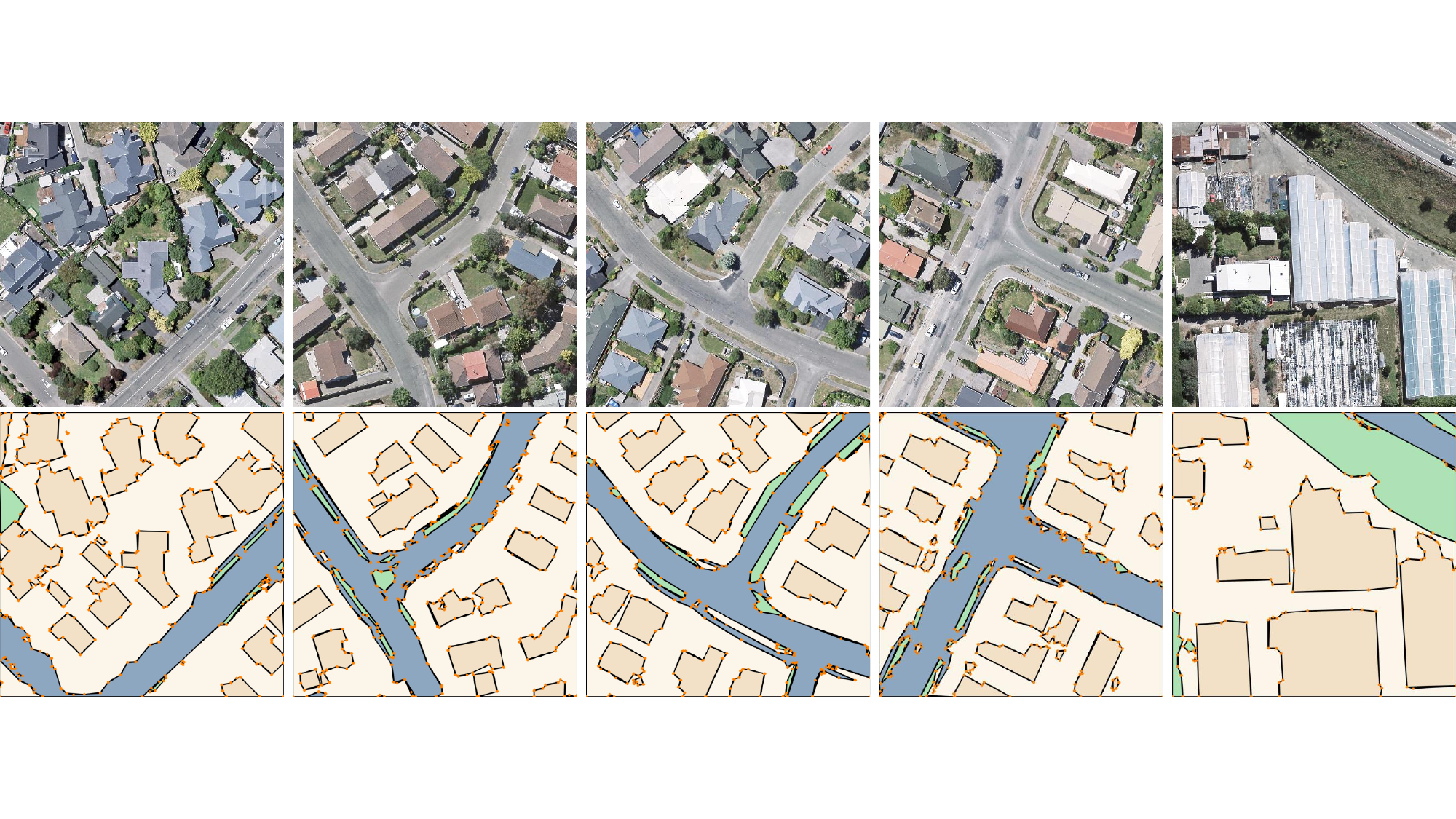}
    \caption{Christchurch (New Zealand)}
\end{subfigure}

\caption{
\textbf{Cross-region qualitative results (Part~2).} Polygons of different semantic classes are shown in distinct colors. Predicted vertices are marked as orange points. Topological inconsistencies, if present, would appear in black (gaps) or red (overlaps). These examples illustrate ACPV-Net’s behavior on imagery from independent national mapping agencies \textbf{without fine-tuning}. 
}
\label{fig:cross_region_vis_part2}
\end{figure*}

Figs~\ref{fig:cross_region_vis_part1} and \ref{fig:cross_region_vis_part2} show representative qualitative results. 
Although polygon regularity naturally degrades under domain shift (different countries, urban layouts, seasons, and illumination), 
ACPV-Net still produces globally coherent vector basemaps with consistent shared boundaries and no visually apparent gaps or overlaps in these examples. 
These observations indicate that, through semantically supervised conditioning, the model learns geometric primitives that are not tied to dataset-specific visual cues, but capture a transferable distribution over aerial-image vertices and cadastral-style partitions.

At the dataset level, these experiments also indicate that, despite its compact spatial extent, Deventer-512 provides sufficiently diverse geometry and visual conditions for training models that generalize reasonably to other high-resolution aerial regions. 
We view Deventer-512 as a compact yet challenging testbed on which future methods can benchmark both in-domain performance and cross-region generalization.

\subsection{Evaluation on the Shanghai dataset}
To further evaluate the generalization ability of ACPV-Net on other public datasets, 
we report additional experiments on the Shanghai dataset used in HiSup \cite{xu2023hisup}. 
We apply the official HiSup predictions and metrics with our comprehensive evaluation.

Table~\ref{tab:R1_shanghai} summarizes the quantitative comparison. 
ACPV-Net achieves competitive or superior performance across most metrics, 
particularly on polygon-level metrics such as CIoU and PoLiS.

\begin{table*}[t]
\centering
\caption{Quantitative results on the Shanghai dataset. 
For baseline methods, AP, AP$_{boundary}$, and CIoU are taken from the HiSup paper, 
while the remaining metrics are computed using the official HiSup predictions with our evaluation protocol.}
\label{tab:R1_shanghai}
\begin{tabular}{lccccccc}
\toprule
Method & AP$\uparrow$ & AP$_{boundary}$$\uparrow$ & CIoU$\uparrow$ & IoU$\uparrow$ & BIoU$\uparrow$ & PoLiS$\downarrow$ & MTA$\downarrow$ \\
\midrule
PolyMapper & 4.3 & 0.8 & 28.6 & -- & -- & -- & -- \\
FFL & 16.3 & 3.9 & 16.4 & -- & -- & -- & -- \\
HiSup & 45.0 & 19.4 & 52.8 & 74.9 & 56.8 & 2.4 & 41.3 \\
Ours & 42.1 & 19.4 & 64.9 & 76.7 & 58.6 & 2.2 & 39.2 \\
\bottomrule
\end{tabular}
\end{table*}

\subsection{TopDiG metrics on Deventer-512}
To enable a more direct comparison with state-of-the-art methods such as TopDiG \cite{yang2023topdig}, 
we additionally report the TopDiG evaluation metrics on Deventer-512. 
These metrics mainly evaluate pixel-level and boundary-based performance, 
which differs from the polygon-level metrics adopted in ACPV. 

Direct comparison on the datasets used in TopDiG is not applicable because TopDiG targets directional graph extraction 
and is primarily evaluated on semantic segmentation datasets without polygonal ground truth, 
whereas ACPV requires vectorized polygons for polygon-level evaluation.

Table~\ref{tab:R2_topdig} presents the macro-averaged results. 
ACPV-Net achieves consistently better performance than TopDiG across the reported metrics.

\begin{table*}[t]
\centering
\caption{TopDiG metrics on Deventer-512 ($\delta = 5$).}
\label{tab:R2_topdig}
\begin{tabular}{lccccccc}
\toprule
Method & PA$_{mask}$ & F1$_{mask}$ & mIoU$_{mask}$ & PA$_{topo}$ & F1$_{topo}$ & mIoU$_{topo}$ & APLS \\
\midrule
TopDiG & 91.4 & 72.1 & 58.0 & 90.4 & 49.6 & 34.6 & 50.7 \\
Ours & 96.4 & 89.8 & 81.6 & 94.2 & 74.9 & 60.4 & 76.2 \\
\bottomrule
\end{tabular}
\end{table*}

\section{Additional Qualitative Visualizations}
\label{sec:sup_more_figs}
To complement the quantitative evaluations in the main paper, we present representative qualitative examples across the principal sources of geometric and visual difficulty in Deventer-512. 
For each scenario category, figures in this section provide side–by–side comparisons between ACPV-Net and the strongest polygonization baselines (HiSup \cite{xu2023hisup}, TopDiG \cite{yang2023topdig}, and GCP \cite{zhang2025global}), illustrating differences in shared-boundary  consistency, vertex localization, and global topology.

\subsection{Multi-Class Geometric Complexity}
\begin{figure*}[t]
\centering
\begin{subfigure}[t]{\linewidth}
    \centering
    \includegraphics[width=\linewidth, trim={0mm 68mm 0mm 68mm}, clip]{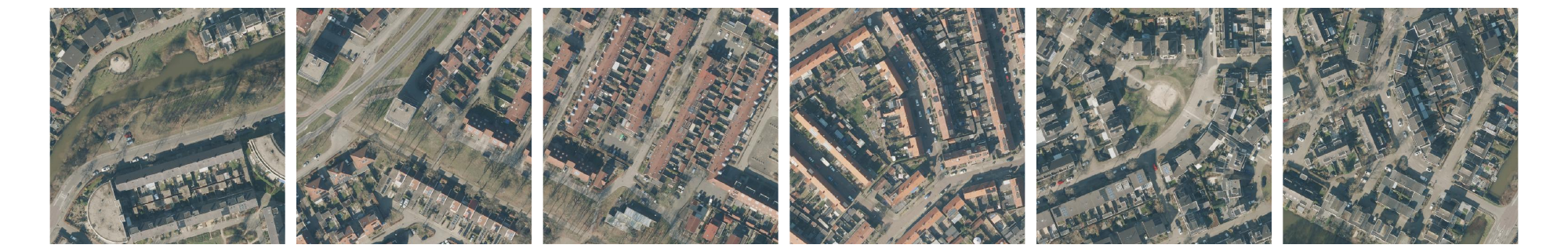}
    \caption{Aerial image}
\end{subfigure}
\begin{subfigure}[t]{\linewidth}
    \centering
    \includegraphics[width=\linewidth, trim={0mm 68mm 0mm 68mm}, clip]{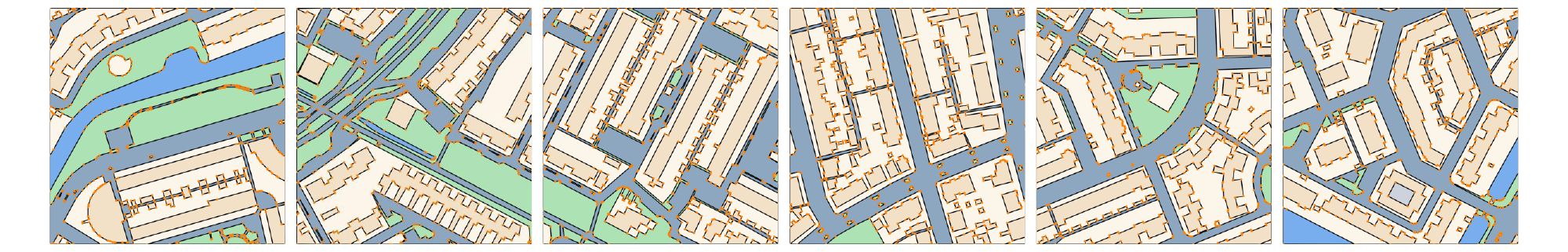}
    \caption{Ground truth}
\end{subfigure}
\begin{subfigure}[t]{\linewidth}
    \centering
    \includegraphics[width=\linewidth, trim={0mm 68mm 0mm 68mm}, clip]{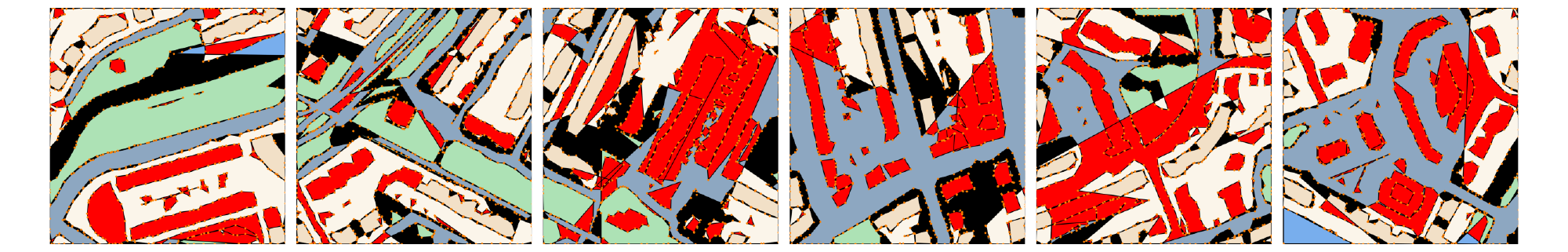}
    \caption{TopDiG}
\end{subfigure}
\begin{subfigure}[t]{\linewidth}
    \centering
    \includegraphics[width=\linewidth, trim={0mm 68mm 0mm 68mm}, clip]{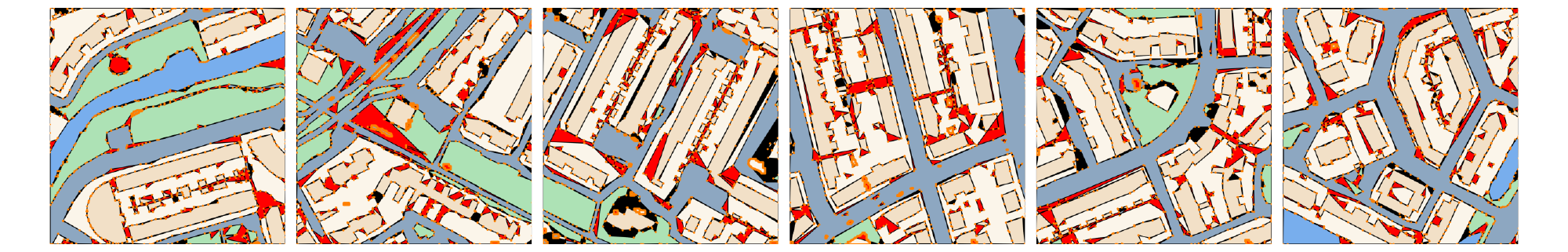}
    \caption{HiSup}
\end{subfigure}
\begin{subfigure}[t]{\linewidth}
    \centering
    \includegraphics[width=\linewidth, trim={0mm 68mm 0mm 68mm}, clip]{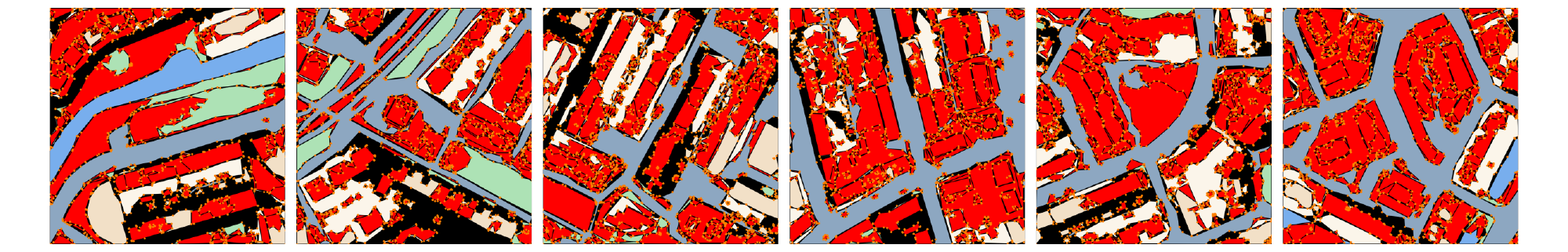}
    \caption{GCP}
\end{subfigure}
\begin{subfigure}[t]{\linewidth}
    \centering
    \includegraphics[width=\linewidth, trim={0mm 68mm 0mm 68mm}, clip]{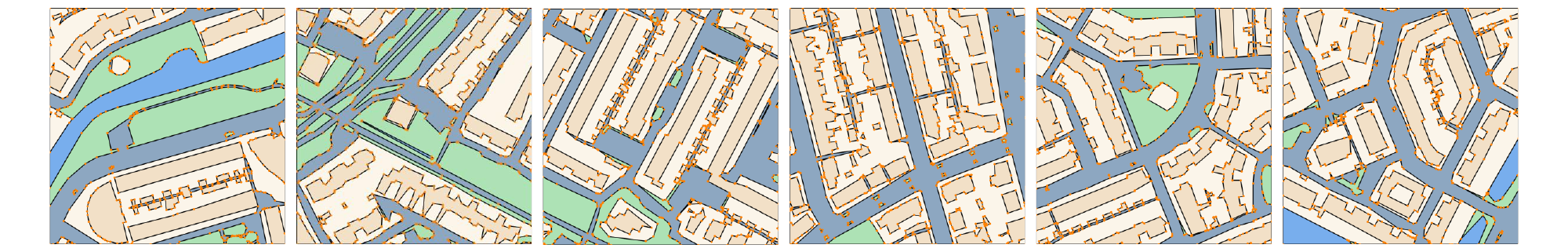}
    \caption{Ours}
\end{subfigure}

\caption{
\textbf{Examples of multi-class geometric complexity.} Polygons of different semantic classes are shown in distinct colors. Predicted vertices are marked as orange points. Topological inconsistencies, if present, would appear in black (gaps) or red (overlaps). 
}
\label{fig:qual_complex_scenes}
\end{figure*}

We first show examples containing dense mixtures of buildings, roads, vegetation, water, and unvegetated regions (Fig.~\ref{fig:qual_complex_scenes}). 
These include large-scale unvegetated parcels with multiple interior holes, nested polygons, and complex multi-class adjacencies. 
ACPV-Net produces a coherent planar partition with clean shared boundaries, whereas stitched single-class baselines frequently introduce gaps, inter-class overlaps, or broken interior rings.

\subsection{Thin and Topologically Fragile Structures}
\begin{figure*}[t]
\centering
\begin{subfigure}[t]{\linewidth}
    \centering
    \includegraphics[width=\linewidth, trim={0mm 68mm 0mm 68mm}, clip]{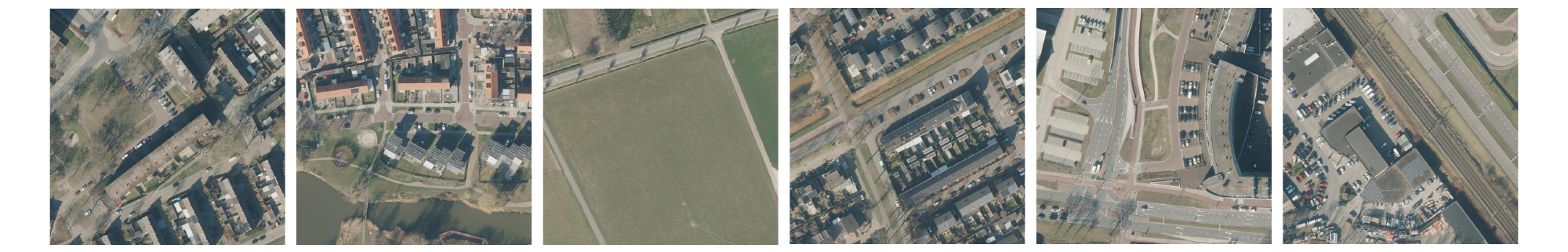}
    \caption{Aerial image}
\end{subfigure}
\begin{subfigure}[t]{\linewidth}
    \centering
    \includegraphics[width=\linewidth, trim={0mm 68mm 0mm 68mm}, clip]{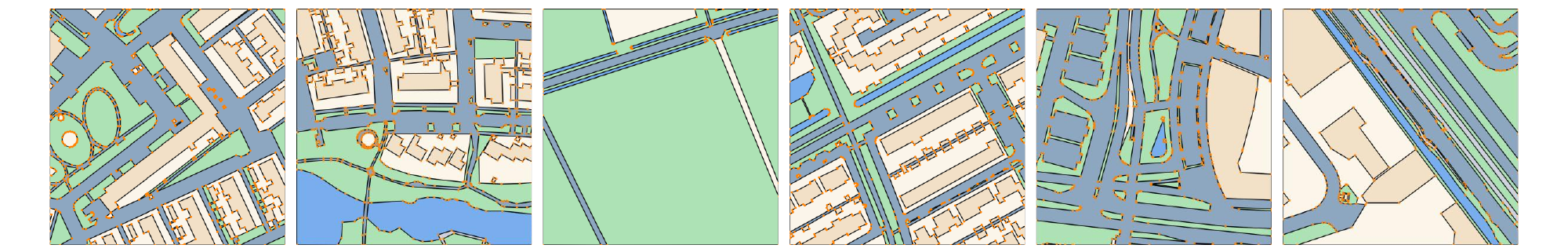}
    \caption{Ground truth}
\end{subfigure}
\begin{subfigure}[t]{\linewidth}
    \centering
    \includegraphics[width=\linewidth, trim={0mm 68mm 0mm 68mm}, clip]{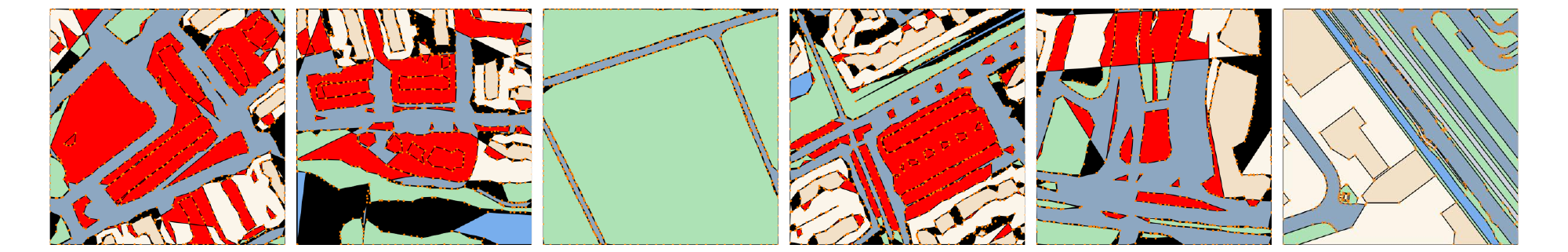}
    \caption{TopDiG}
\end{subfigure}
\begin{subfigure}[t]{\linewidth}
    \centering
    \includegraphics[width=\linewidth, trim={0mm 68mm 0mm 68mm}, clip]{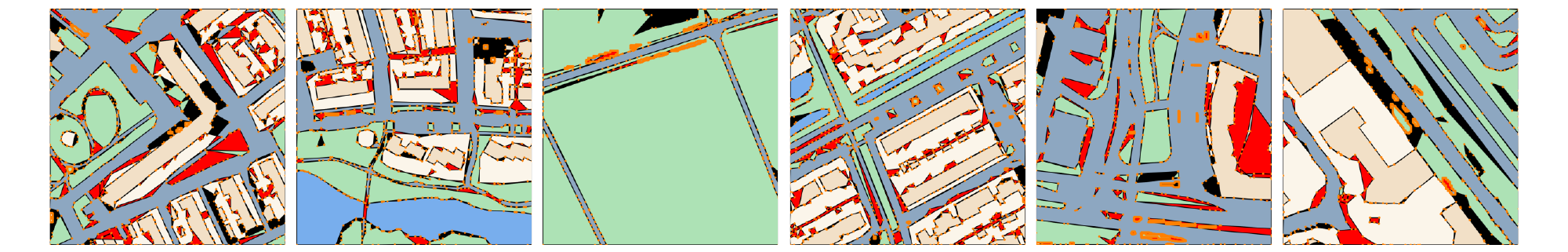}
    \caption{HiSup}
\end{subfigure}
\begin{subfigure}[t]{\linewidth}
    \centering
    \includegraphics[width=\linewidth, trim={0mm 68mm 0mm 68mm}, clip]{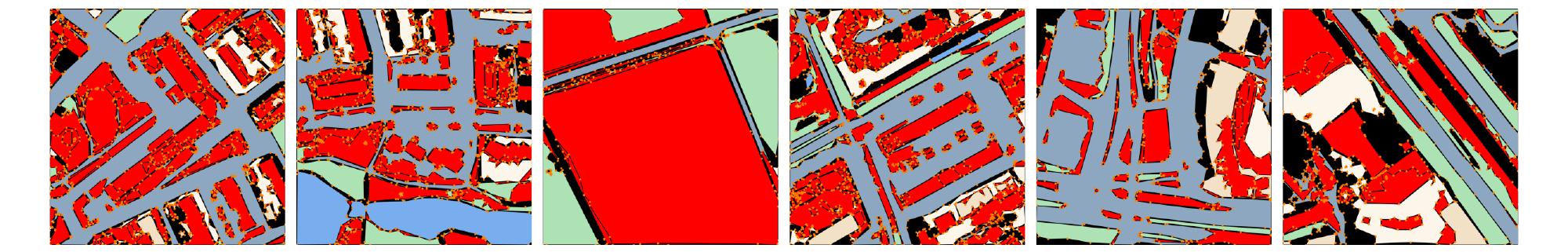}
    \caption{GCP}
\end{subfigure}
\begin{subfigure}[t]{\linewidth}
    \centering
    \includegraphics[width=\linewidth, trim={0mm 68mm 0mm 68mm}, clip]{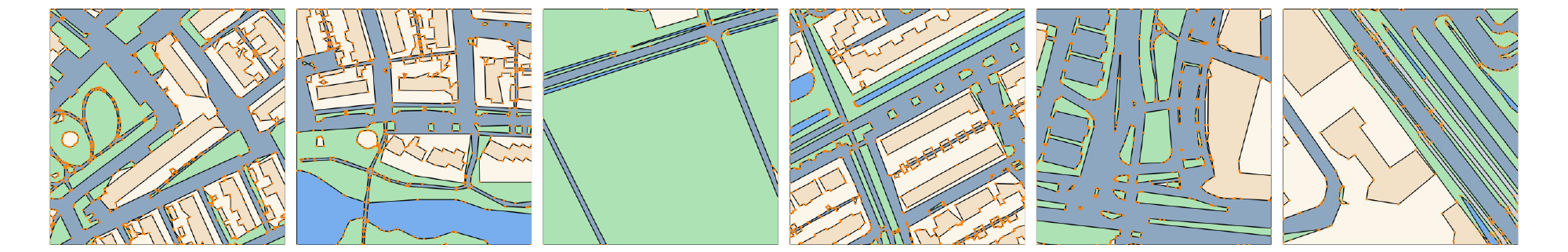}
    \caption{Ours}
\end{subfigure}

\caption{
\textbf{Examples of thin and topologically fragile Structures.} Polygons of different semantic classes are shown in distinct colors. Predicted vertices are marked as orange points. Topological inconsistencies, if present, would appear in black (gaps) or red (overlaps).
}
\label{fig:qual_thin_structures}
\end{figure*}

Fig.~\ref{fig:qual_thin_structures} presents narrow and fragile structures such as thin roads, elongated water bodies, vegetation strips, and small garden beds. 
These geometries are extremely sensitive to the boundary localization errors. 
ACPV-Net preserves connectivity and topological integrity, while baselines commonly exhibit breakage, merging, or self-intersecting segments in these cases.

\subsection{Challenging Visual Conditions (Successes and Limitations)}
\begin{figure*}[t]
\centering
\begin{subfigure}[t]{\linewidth}
    \centering
    \includegraphics[width=\linewidth, trim={0mm 68mm 0mm 68mm}, clip]{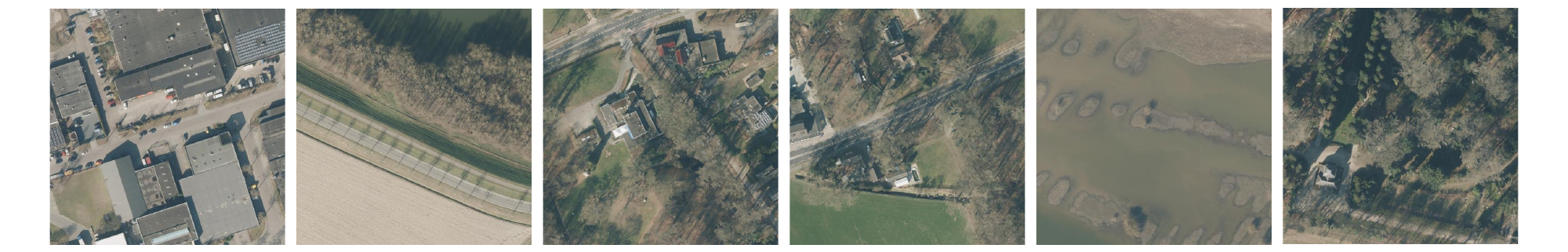}
    \caption{Aerial image}
\end{subfigure}
\begin{subfigure}[t]{\linewidth}
    \centering
    \includegraphics[width=\linewidth, trim={0mm 68mm 0mm 68mm}, clip]{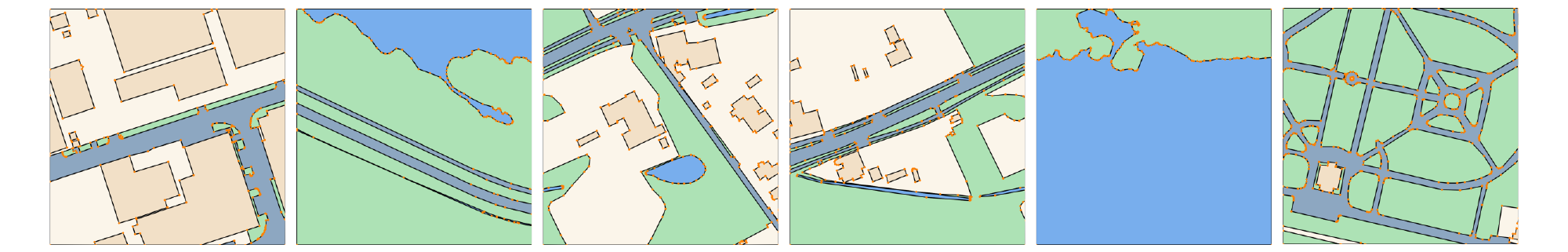}
    \caption{Ground truth}
\end{subfigure}
\begin{subfigure}[t]{\linewidth}
    \centering
    \includegraphics[width=\linewidth, trim={0mm 68mm 0mm 68mm}, clip]{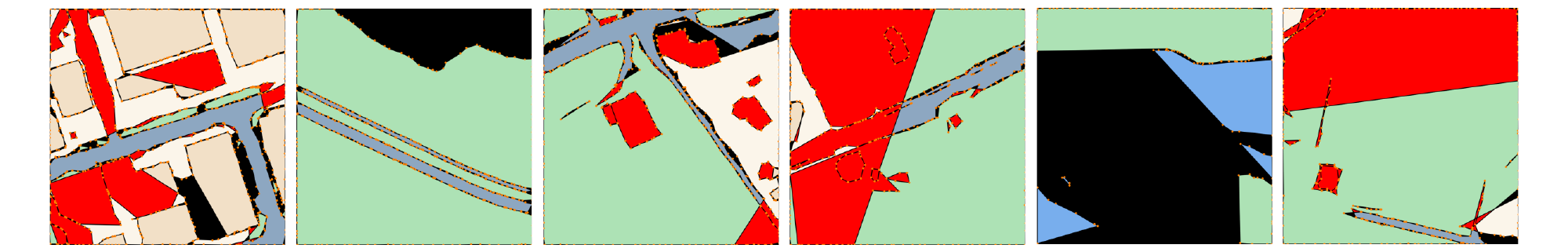}
    \caption{TopDiG}
\end{subfigure}
\begin{subfigure}[t]{\linewidth}
    \centering
    \includegraphics[width=\linewidth, trim={0mm 68mm 0mm 68mm}, clip]{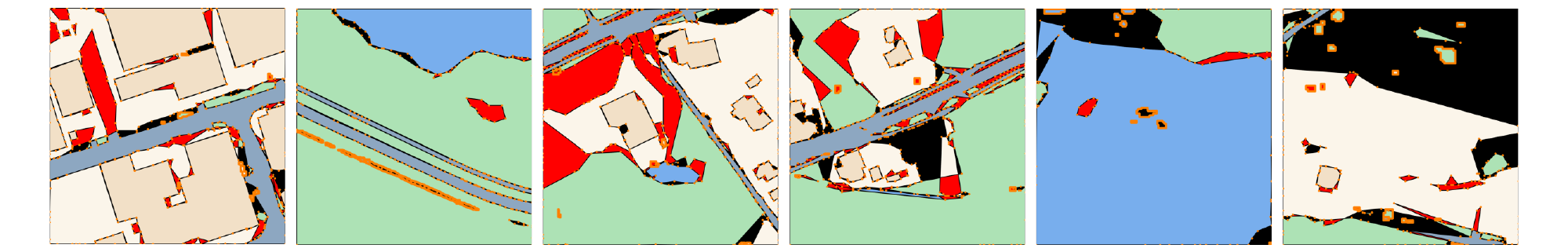}
    \caption{HiSup}
\end{subfigure}
\begin{subfigure}[t]{\linewidth}
    \centering
    \includegraphics[width=\linewidth, trim={0mm 68mm 0mm 68mm}, clip]{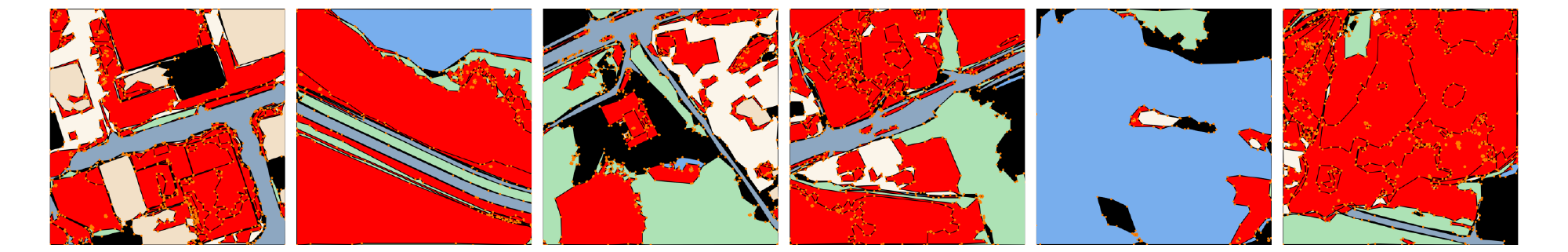}
    \caption{GCP}
\end{subfigure}
\begin{subfigure}[t]{\linewidth}
    \centering
    \includegraphics[width=\linewidth, trim={0mm 68mm 0mm 68mm}, clip]{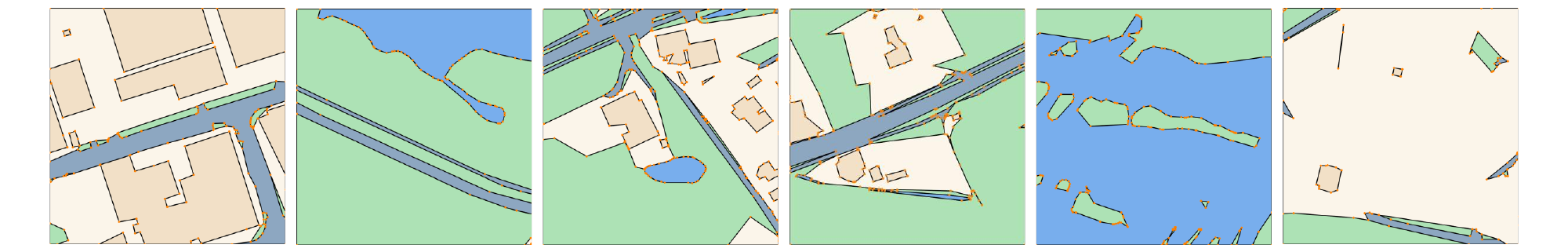}
    \caption{Ours}
\end{subfigure}

\caption{
\textbf{Examples of challenging visual conditions.} Polygons of different semantic classes are shown in distinct colors. Predicted vertices are marked as orange points. Topological inconsistencies, if present, would appear in black (gaps) or red (overlaps).
}
\label{fig:qual_visual_conditions}
\end{figure*}

In Fig.~\ref{fig:qual_visual_conditions}, we collect examples with strong shadows, weak or missing texture, reflective surfaces, dense vegetation boundaries, and cadastral–visual mismatches where the true boundary is not visually implied. 
ACPV-Net remains robust under most of these adverse conditions, although large cadastral–visual discrepancies can still degrade polygon regularity. 
These visualizations highlight both the strengths of semantically supervised conditioning and the remaining failure modes inherent to ambiguous imagery.

\noindent\textbf{Limitation.} As shown in Fig.~\ref{fig:qual_visual_conditions}, strong occlusion and cadastral-visual mismatch can lead to suboptimal performance. Also, the semantically ambiguous boundaries as shown in Sec.~\ref{sec:sup_challengingvisual} make the polygon vectorization challenging. In future work, we will seek uncertainty quantification to outline such cases, in order to guide the correction with minimum human effort. 

\clearpage
\balance
{
    \small
    \bibliographystyle{ieeenat_fullname}
    \bibliography{main}
}